\documentclass[review]{fcs}
\usepackage{microtype}
\usepackage{mathrsfs}
\usepackage{longtable}
\usepackage[english]{babel}
\usepackage{threeparttable}
\usepackage{pifont}
\usepackage{wasysym}
\usepackage{times}
\usepackage[utf8]{inputenc}
\usepackage{inconsolata}
\usepackage{hyperref}
\usepackage{dcolumn}
\usepackage{tabularx}
\usepackage{colortbl}
\usepackage{url}
\usepackage{graphicx}
\usepackage{color}
\usepackage{enumitem}

\usepackage[normalem]{ulem}
\usepackage{makecell}
\usepackage{pifont}
\usepackage[edges]{forest}
\usepackage{amsmath,amsthm,amsfonts,amssymb,bm}
\usepackage{microtype}
\usepackage{xspace,mfirstuc,tabulary}
\usepackage{booktabs}
\usepackage{pifont}
\usepackage{multirow,booktabs, hhline}
\usepackage{arydshln}
\usepackage{bbm}
\usepackage{subfigure}
\usepackage{makecell}
\usepackage{CJKutf8}
\usepackage{cleveref}
\usepackage{listings}
\usepackage{titlesec}
\crefname{section}{§}{§§}
\Crefname{section}{§}{§§}
\usepackage{wrapfig}
\usepackage{lipsum}
\usepackage{pgfplots}
\usepackage{tikz}
\usetikzlibrary{calc}
\usepackage{booktabs}
\usepackage{array}
\usepackage{makecell}
\usepackage{pifont}
\usepackage{tcolorbox} 
\usepackage{setspace}

\newenvironment{itemize*}%
 {\leftmargini=20pt\begin{itemize}%
  \setlength{\itemsep}{3pt}%
  \setlength{\parskip}{0pt}%
  }%
 {\end{itemize}}
\newenvironment{enumerate*}%
 {\begin{enumerate}%
  \setlength{\itemsep}{0pt}%
  \setlength{\parskip}{0pt}}%
 {\end{enumerate}}
 
\definecolor{lightergray}{RGB}{230,230,230}
\definecolor{DarkGreen}{RGB}{30,130,30}
\newcommand{\cmark}{\textcolor{DarkGreen}{\ding{51}}}
\newcommand{\xmark}{\textcolor{red}{\ding{55}}}%

\definecolor{LightCyan}{rgb}{0.88,1,1}
\newcommand{\paratitle}[1]{\vspace{1.5ex}\noindent\textbf{#1}}
\newcommand{\ie}{\emph{i.e.,}\xspace}

\newcommand{\eg}{\emph{e.g.,}\xspace}

\newcommand{\etc}{\emph{etc}}

\title{Tool Learning with Large Language Models: A Survey}
\author[1]{Changle QU}
\author[1]{Sunhao DAI}
\author[2]{Xiaochi WEI}
\author[3]{Hengyi CAI}
\author[2]{Shuaiqiang WANG}
\author[2]{Dawei YIN}
\author*[1]{Jun XU}
\author[1]{Ji-Rong WEN}
\address[1]{Gaoling School of Artificial Intelligence, Renmin University of China, Beijing, 100872, China}
\address[2]{Baidu Inc., Beijing 100193, China}
\address[3]{Institute of Computing Technology, Chinese Academy of Sciences, Beijing 100864, China}
\fcssetup{
  received       = {month dd, yyyy},
  accepted       = {month dd, yyyy},
  corr-email     = {junxu@ruc.edu.cn},
}
\begin{abstract}
Recently, tool learning with large language models~(LLMs) has emerged as a promising paradigm for augmenting the capabilities of LLMs to tackle highly complex problems. 
Despite growing attention and rapid advancements in this field, the existing literature remains fragmented and lacks systematic organization, posing barriers to entry for newcomers. 
This gap motivates us to conduct a comprehensive survey of existing works on tool learning with LLMs.
In this survey, we focus on reviewing existing literature from the two primary aspects (1) why tool learning is beneficial and (2) how tool learning is implemented, enabling a comprehensive understanding of tool learning with LLMs.
We first explore the ``why'' by reviewing both the benefits of tool integration and the inherent benefits of the tool learning paradigm from six specific aspects. In terms of ``how'', we systematically review the literature according to a taxonomy of four key stages in the tool learning workflow: task planning, tool selection, tool calling, and response generation. 
Additionally, we provide a detailed summary of existing benchmarks and evaluation methods, categorizing them according to their relevance to different stages. 
Finally, we discuss current challenges and outline potential future directions, aiming to inspire both researchers and industrial developers to further explore this emerging and promising area.
\end{abstract}

\begin{document}

\section{Introduction}
\label{sec:intro}
\begin{flushright}
\rightskip=3.6cm\textit{``Sharp tools make good work.''} \\
\vspace{.2em}
\rightskip=.0cm---\emph{The Analects: Wei Ling Gong}
\end{flushright}

Throughout history, humanity has continually sought innovation, utilizing increasingly sophisticated tools to boost efficiency and enhance capabilities~\cite{washburn1960tools,gibson1993tools}. 
These tools, extending both our intellect and physicality, have been crucial in driving social and cultural evolution~\cite{von1995cognitive}. 
From primitive stone tools to advanced machinery, this progression has expanded our potential beyond natural limits, enabling more complex and efficient task management~\cite{shumaker2011animal}. 

Today, we are experiencing a new technological renaissance, driven by breakthroughs in artificial intelligence, especially through the development of large language models (LLMs). 
Pioneering models such as ChatGPT~\cite{achiam2023gpt} have demonstrated remarkable capabilities, marking significant progress in a range of natural language processing (NLP) tasks, including summarization~\cite{el2021automatic,zhang2024benchmarking}, machine translation~\cite{zhang2023prompting,feng2024improving}, question answering~\cite{yang2018hotpotqa,kwiatkowski2019natural}, \etc.
However, despite their impressive capabilities, LLMs often struggle with complex computations and delivering accurate, timely information due to their reliance on fixed and parametric knowledge~\cite{mallen2022not,vu2023freshllms}. 
This inherent limitation frequently results in responses that are plausible yet factually incorrect or outdated (often referred to as hallucination)~\cite{ji2023survey,zhang2023siren}, posing significant risks and misleading users.

\begin{figure*}[t]
\centering
	\includegraphics[width=\linewidth]{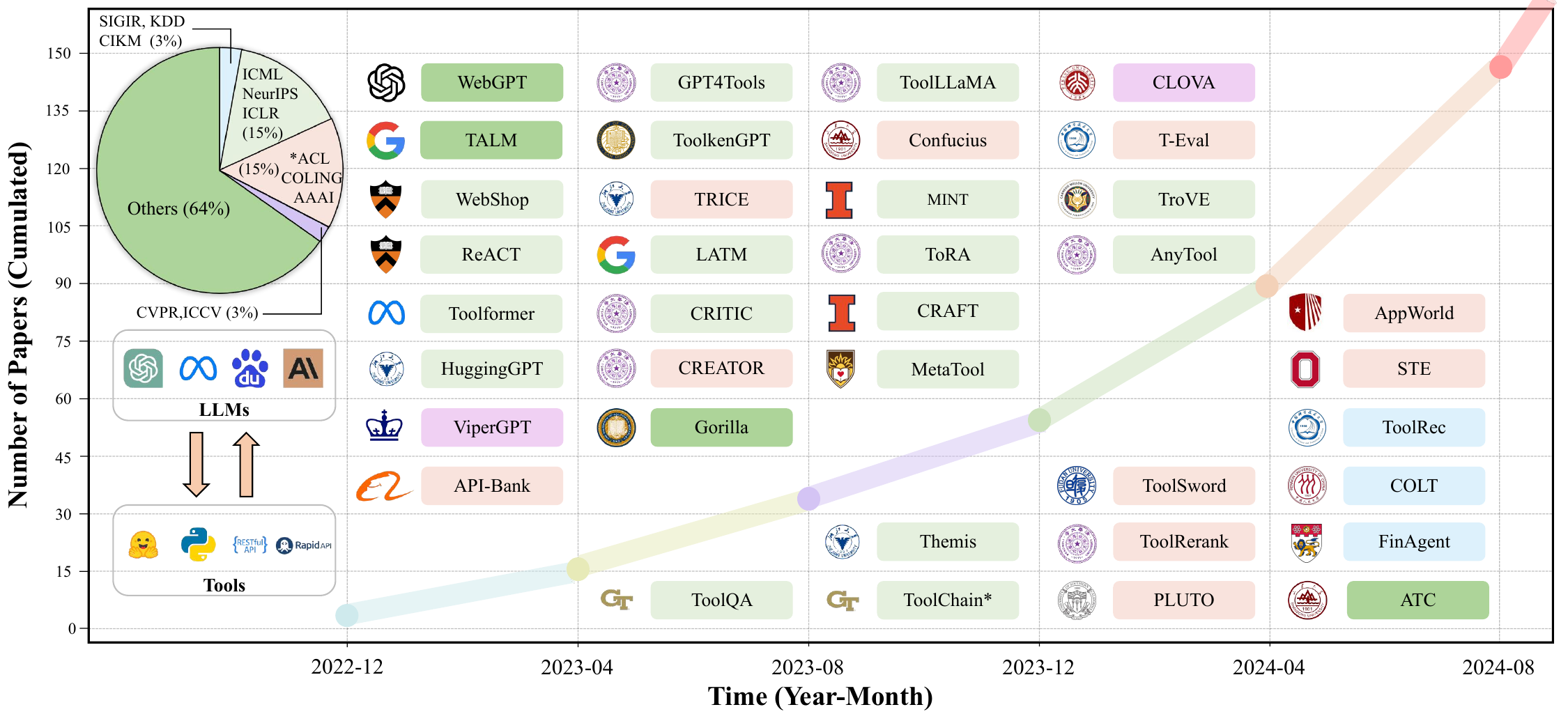}
\caption{An illustration of the development trajectory of tool learning. We present the statistics of papers with the publication year and venue, with each venue uniquely represented by a distinct color. For each time period, we have selected a range of representative landmark studies that have significantly contributed to the field.~(Note that we use the institution of the first author as the representing institution in the figure.)}
        \label{fig:growth}
\end{figure*}

With the continuous enhancement of LLMs capabilities, it is expected that LLMs will become proficient in using tools to solve complex problems as human~\cite{qin2023tool}, a concept known as tool learning with LLMs.
Tool learning emerges as a promising solution to mitigate these limitations of LLMs by enabling dynamic interaction with external tools~\cite{schick2024toolformer,qin2023toolllm, tang2023toolalpaca,wang2024empowering}. 
This approach not only enhances problem-solving capabilities of LLMs but also broadens their functional scope~\cite{yao2022webshop,lazaridou2022internet,lu2023gear}. 
For instance, LLMs can perform complex calculations using a calculator tool, access real-time weather updates through weather APIs, and execute programming code via interpreters~\cite{pan-etal-2023-fact,wang2023mint}. 
This integration significantly improves their response accuracy to user queries, facilitating more effective and reliable user interactions. 
As this field continues to evolve, tool-augmented LLMs are expected to play a pivotal role in the future of NLP~\cite{parisi2022talm, karpas2022mrkl}, offering more versatile and adaptable solutions~\cite{nakano2021webgpt,suris2023vipergpt}.

As shown in Figure~\ref{fig:growth}, the past year has witnessed a rapid surge in research efforts on tool learning concurrent with the rise of LLMs. 
Notably, in practical applications, GPT-4~\cite{achiam2023gpt} addresses its knowledge limitations and augments its capabilities by calling on plugins, ultimately integrating the returned results of plugins with its internal knowledge to generate better responses for users.
Within the research community, much effort has been made in exploring how to evaluate the tool learning capabilities of LLMs~\cite{li2023api,huang2023metatool,chen2023t} and how to enhance it to strengthen the capabilities of LLMs~\cite{xu2023tool,Gao2023ConfuciusIT,zhao2024let}.
Given the increasing attention and rapid development of tool learning with LLMs, it is essential to systematically review the most recent advancements and challenges, so as to benefit researchers and industrial developers in understanding the current progress and inspire more future work in this area.

In this survey, we conduct a systematic exploration of existing studies in two primary dimensions: (1) \textbf{why tool learning} is beneficial and (2) \textbf{how tool learning} is implemented. Specifically, the ``why tool learning'' dimension examines both the advantages of tool integration and the inherent benefits of the tool learning paradigm, while the ``how tool learning'' dimension details the four stages of the entire tool learning workflow: task planning, tool selection, tool calling, and response generation. These dimensions are foundational to understanding tool learning with LLMs. Moreover, we provide a systematic summary of existing benchmarks and evaluation methods, classifying them based on their focus across different stages. Finally, we discuss the current challenges and propose future directions, offering critical insights to facilitate the development of this promising and burgeoning research area. We also maintain a GitHub repository to continually keep track of the relevant papers and resources in this rising area at \textcolor{blue}{\url{https://github.com/quchangle1/LLM-Tool-Survey}}.

It is worth noting that while other surveys provide comprehensive overviews of techniques and methods used by LLMs~\cite{zhao2023survey}, applications in planning~\cite{huang2024understanding}, reasoning~\cite{qiao2022reasoning,sun2023survey}, agents~\cite{wang2024survey,sumers2024cognitive,xi2023rise}, and retrieval-augmented generation~\cite{gao2023retrieval,zhao2024retrieval}, they often mention tools or tool learning but do not extensively explore this aspect. 
Compared with them, our survey provides a focused and detailed analysis of tool learning with LLMs, especially elucidating the dual aspects of \textbf{why} tool learning is essential for LLMs and \textbf{how} tool learning can be systematically implemented.
Through these two principle aspects, we offer an up-to-date and comprehensive review of tool learning with LLMs. 
Meanwhile, we also acknowledge the foundational contributions of earlier perspective papers like those by Mialon et al. (2023)~\cite{mialon2023augmented} and Qin et al. (2023)~\cite{qin2023tool}, which initially highlighted the promising opportunities that tools present to enhance LLMs capabilities. 
Since the field has seen rapid growth with many new studies emerging, our survey provides a broader introduction to these latest developments.
Additionally, a more recent survey \cite{wang2024tools} discusses various tooling scenarios and approaches employed in language models, serving as an excellent supplement to our comprehensive review.

\definecolor{hidden-draw}{RGB}{205, 44, 36}
\definecolor{hidden-orange}{RGB}{243,202,120}
\definecolor{hidden-blue}{RGB}{194,232,247}
\definecolor{hidden-yellow}{RGB}{242,244,193}
\definecolor{tree-level-1}{RGB}{245,20,85}
\definecolor{tree-level-2}{RGB}{246,86,118}
\definecolor{tree-level-3}{RGB}{248,177,193}
\definecolor{tree-leaf}{RGB}{176,230,198}
\definecolor{Self}{RGB}{255,0,128}
\definecolor{Ensemble}{RGB}{0,127,255}
\definecolor{Iterative}{RGB}{153,51,255}
\definecolor{exemplar1}{RGB}{136,98,148}
\definecolor{exemplar2}{RGB}{148,210,242}
\definecolor{knowledge1}{RGB}{249,219,152}
\definecolor{knowledge2}{RGB}{255,245,220}


\tikzstyle{my-box}=[
    rectangle,
    draw=hidden-draw,
    rounded corners,
    text opacity=1,
    minimum height=1.5em,
    minimum width=5em,
    inner sep=2pt,
    align=center,
    fill opacity=.5,
]
\tikzstyle{leaf}=[my-box, minimum height=1.5em,
    fill=hidden-orange!60, text=black, align=left,font=\scriptsize,
    inner xsep=2pt,
    inner ysep=4pt,
]
\begin{figure*}[!t]
    \centering
    \resizebox{\textwidth}{!}{
        \begin{forest}
            forked edges,
            for tree={
                grow=east,
                reversed=true,
                anchor=base west,
                parent anchor=east,
                child anchor=west,
                base=left,
                font=\small,
                rectangle,
                draw=hidden-draw,
                rounded corners,
                align=left,
                minimum width=4em,
                edge+={darkgray, line width=1pt},
                s sep=3pt,
                inner xsep=2pt,
                inner ysep=3pt,
                ver/.style={rotate=90, child anchor=north, parent anchor=south, anchor=center},
            },
            where level=1{text width=5.0em,font=\scriptsize,}{},
            where level=2{text width=6.3em,font=\scriptsize,}{},
            where level=3{text width=7.0em,font=\scriptsize,}{},
            where level=4{text width=6.1em,font=\scriptsize,}{},
            [
                Tool~ Learning~ with~ LLMs, ver
                [
                    Why Tool \\ Learning~(\S \ref{sec:why})
                    [
                        Benefits of Tool \\  Integration
                        [
                            Knowledge \\Acquisition~(\S \ref{subsec:knowledge Acquisition})
                            [
                                (1)~Search Engine~\cite{komeili2021internet,nakano2021webgpt,zhang2023toolcoder,shi2023replug,schick2024toolformer,paranjape2023art,gou2023critic}{, }
                                \\
                                (2)~Database and Knowledge Graph~\cite{thoppilan2022lamda,patil2023gorilla,hao2024toolkengpt,zhuang2024toolqa,zhang2024syntax,gu2024middleware}{, }
                                \\
                                (3)~Weather or Map~\cite{xu2023tool,tang2023toolalpaca,qin2023tool,huang2023metatool}{.}, leaf, text width=31.9em
                            ]
                        ]
                        [
                            Expertise\\Enhancement~(\S \ref{subsec:expertise})
                            [
                                (1)~Mathematical Tools~\cite{cobbe2021training,karpas2022mrkl,shao2022chaining,kadlvcik2023calc,he2023solving,zhang2024evaluating,gou2023tora,das2024mathsensei,veerendranath2024calccmu,bulusu2024mathviz}{, }\\
                                (2)~Python Interpreter~\cite{gao2023pal,chen2022program,pan-etal-2023-fact,lu2024chameleon,wang2023leti,wang2023mint,wu2024structureaware,zhang2024codeagent}{, }\\
                                (3)~Others~\cite{inaba-etal-2023-multitool,m2024augmenting,ramos2024review,jin2024genegpt,theuma2024equipping,gao2024simulating,zhang2024multimodal,jin2024agentmd,li2024mmedagent,zhao2024let}{.}
                                , leaf, text width=31.9em
                            ]
                        ]
                        [
                            Automation and \\Efficiency~(\S \ref{subsec:action})
                            [
                                (1)~Schedule Tools~\cite{schick2024toolformer}{, } \\
                                (2)~Set Reminders~\cite{zhuang2024toolqa}{, }\\
                                (3)~Filter Emails~\cite{qin2023toolllm}{, } \\
                                (4)~Project Management~\cite{qin2023toolllm}{, }\\
                                (5)~Online Shopping Assistants~\cite{yao2022webshop}{.}
                                , leaf, text width=31.9em
                            ]
                        ]
                        [
                            Interaction \\Enhancement~(\S \ref{subsec:multi-modal})
                            [
                                (1)~Multi-modal Tools~\cite{suris2023vipergpt,yang2023mm,liu2023internchat,gao2023assistgpt,gao2023clova,zhao2024diffagent,ma2024m,wang2024tool}{, }\\
                                (2)~Machine Translator~\cite{qin2023tool,schick2024toolformer}{, } \\
                                (3)~Natural Language Processing Tools~\cite{qin2023toolllm,shen2024hugginggpt,lyu2023gitagent}{.}
                                , leaf, text width=31.9em
                            ]
                        ]
                    ]
                    [
                        Benefits of Tool \\Learning
                        [
                            (1)~Enhanced Interpretability and User Trust~(\S \ref{subsec:interpretability}){, }
                            (2)~Improved Robustness and Adaptability~(\S \ref{subsec:robustness}){.}
                            , leaf, text width=40.5em
                        ]
                    ]
                ]
                [
                    How Tool \\ Learning~(\S \ref{sec:how tool learning})
                    [
                        Overview of the \\Paradigm~(\S \ref{subsec:overview})
                        [
                            Four Stages of Tool Learning:~(1)~Task Planning{, }(2)~Tool Selection{, }(3)~Tool Calling{, }(4)~Response Generation{.}\\
                            Two Paradigms of Tool Learning:~(1)~Single Invocation Tool Learning{, }(2)~Iterative Invocation Tool Learning{.}
                            , leaf, text width=40.5em
                        ]
                    ]
                    [
                        Task Planning\\~(\S \ref{subsec:planning})
                        [
                            Tuning-free \\Methods~(\S\ref{subsubsec:planning_free})
                            [
                                CoT~\cite{wei2022chain}{, }ReACT~\cite{yao2022react}{, }ART~\cite{paranjape2023art}{, }RestGPT~\cite{song2023restgpt}{, }HuggingGPT~\cite{shen2024hugginggpt}{, }TPTU~\cite{ruan2023tptu}{, }\\ToolChain*\cite{zhuang2023toolchain}{, }ControlLLM~\cite{liu2023controlllm}{, }Attention Buckets~\cite{chen2023fortify}{, }PLUTO~\cite{huang2024planedit}{, }ATC~\cite{shi2024chain}{, }\\{SGC~\cite{wu2024can}{, }Sum2Act~\cite{liu2024summary}{, }BTP~\cite{zheng2024budget}{, }DRAFT~\cite{qu2024from}{.}}
                                , leaf, text width=31.9em
                            ]
                        ]
                        [
                            Tuning-based \\Methods~(\S\ref{subsubsec:planning_tune})
                            [
                                Toolformer~\cite{schick2024toolformer}{, }TaskMatrix.AI~\cite{liang2024taskmatrix}{, }Toolink~\cite{qian2023toolink}{, }TPTU-v2~\cite{kong2023tptuv2}{, }$\alpha$-UMi~\cite{shen2024small}{, }\\COA~\cite{gao2024efficient}{, }DEER~\cite{gui2024look}{, }OpenAGI~\cite{ge2024openagi}{, }SOAY~\cite{wang2024solution}{, }TP-LLAMA~\cite{chen2024advancing}{, }{APIGen~\cite{liu2024apigen}}{.}
                                , leaf, text width=31.9em
                            ]
                        ]
                    ]
                    [
                        Tool Selection\\~(\S \ref{subsec:selection})
                        [
                            Retriever-based Tool \\Selection~(\S\ref{subsec:retriever})
                            [
                                TF-IDF~\cite{sparck1972statistical}{, }BM25~\cite{robertson2009probabilistic}{, }Sentence-Bert~\cite{reimers2019sentence}{, }ANCE~\cite{xiong2020approximate}{, }TAS-B~\cite{hofstatter2021efficiently}{, }Contriever~\cite{izacard2021unsupervised}{, }\\coCondensor~\cite{gao2021unsupervised}{, }CRAFT~\cite{yuan2023craft}{, }ProTIP~\cite{anantha2023protip}{, }ToolRerank~\cite{zheng2024toolrerank}{, }COLT~\cite{qu2024towards}{.}
                            , leaf, text width=31.9em
                            ]
                        ]
                        [
                            LLM-based Tool \\Selection~(\S\ref{subsec:llm selection})
                            [
                                COT~\cite{wei2022chain}{, }ReACT~\cite{yao2022react}{, }ToolLLaMA~\cite{qin2023toolllm}{, }Confucius~\cite{Gao2023ConfuciusIT}{, }ToolBench~\cite{xu2023tool}{, }RestGPT~\cite{song2023restgpt}{, }\\HuggingGPT~\cite{shen2024hugginggpt}{, }{ChatCoT~\cite{chen-etal-2023-chatcot}}{, }ToolNet~\cite{liu2024toolnet}{, }ToolVerifier~\cite{mekala2024toolverifier}{, }TRICE~\cite{qiao2023making}{, }\\AnyTool~\cite{du2024anytool}{, }GeckOpt~\cite{fore2024geckopt}{.}
                            , leaf, text width=31.9em
                            ]
                        ]
                    ]
                    [
                        Tool Calling~(\S\ref{subsec:tool calling})
                        [
                            Tuning-free \\Methods~(\S\ref{subsubsec:calling_free})
                            [
                                RestGPT~\cite{song2023restgpt}{, }Reverse Chain~\cite{zhang2023reverse}{, }ControlLLM~\cite{liu2023controlllm}{, }EasyTool ~\cite{yuan2024easytool}{, }ToolNet~\cite{liu2024toolnet}{, }\\ConAgents~\cite{shi2024learning}{.}
                                , leaf, text width=31.9em
                            ]
                        ]
                        [
                            Tuning-based \\Methods~(\S\ref{subsubsec:call_tune})
                            [
                                Gorilla~\cite{patil2023gorilla}{, }GPT4Tools~\cite{yang2024gpt4tools}{, }ToolkenGPT\cite{hao2024toolkengpt}{, }Themis~\cite{li2023tool}{, }STE~\cite{wang2024llms}{, }\\ToolVerifier~\cite{mekala2024toolverifier}{, }TRICE~\cite{qiao2023making}{.}
                                , leaf, text width=31.9em
                            ]
                        ]
                    ]
                    [
                        Response \\Generation~(\S\ref{subsec:response generation})
                        [
                            Direct Insertion \\Methods~(\S\ref{subsubsec:response_direct})
                            [
                                TALM~\cite{parisi2022talm}{, }Toolformer~\cite{schick2024toolformer}{, }ToolkenGPT~\cite{hao2024toolkengpt}{.}, leaf, text width=31.9em
                            ]
                        ]
                        [
                            Information \\Integration Methods\\~(\S\ref{subsubsec:response_integration})
                            [
                                RestGPT~\cite{song2023restgpt}{, }ToolLLaMA~\cite{qin2023toolllm}{, }ReCOMP~\cite{xu2023recomp}{, }ConAgents~\cite{shi2024learning}{.}, leaf, text width=31.9em
                            ]
                        ]
                    ]
                ]
                [
                    Benchmarks \\Toolkits and \\Evaluation~(\S \ref{sec:benchmarks and evaluation})
                    [
                        Benchmarks~(\S\ref{subsec:benchmarks})
                        [
                            General Benchmarks
                            [
                                ToolBench2~\cite{qin2023toolllm}{, }APIBench~\cite{patil2023gorilla}{, }ToolBench1~\cite{xu2023tool}{, }ToolAlpaca~\cite{tang2023toolalpaca}{, }RestBench~\cite{song2023restgpt}{, }\\ API-Bank~\cite{li2023api}{, }T-Eval~\cite{chen2023t}{, }MetaTool~\cite{huang2023metatool}{, }ToolEyes~\cite{ye2024tooleyes}{, }UltraTool~\cite{huang2024planning}{, }Seal-Tools~\cite{wu2024seal}{, }\\API-BLEND~\cite{basu2024api}{, }ShortcutsBench~\cite{shen2024shortcutsbench}{, }GTA~\cite{wang2024gta}{, }WTU-Eval~\cite{ning2024wtu}{, }AppWorld~\cite{trivedi2024appworld}{.}, leaf, text width=31.9em
                            ]
                        ]
                        [
                            Other Benchmarks
                            [
                                ToolQA~\cite{zhuang2024toolqa}{, }ToolEmu~\cite{ruan2023identifying}{, }ToolTalk~\cite{farn2023tooltalk}{, }m\&m's~\cite{ma2024m}{, }VIoT~\cite{zhong2023viotgpt}{, }RoTBench~\cite{ye2024rotbench}{, }\\MLLM-Tool~\cite{wang2024tool}{, }ToolSword~\cite{ye2024toolsword}{, }SCITOOLBENCH~\cite{ma2024sciagent}{, }StableToolBench~\cite{guo2024stabletoolbench}{, }\\InjecAgent~\cite{zhan2024injecagent}{, }CToolEval~\cite{guo2024ctooleval}{, }ToolLens~\cite{qu2024towards}{, }SoAyBench\cite{wang2024solution}{, }ToolSandbox~\cite{lu2024toolsandbox}{.}, leaf, text width=31.9em
                            ]
                        ]
                    ]
                    [
                        Toolkits~(\S\ref{subsubsec:toolkit})
                    ]
                    [
                        Evaluation~(\S\ref{subsec:evaluation})
                        [
                            Task Planning:~Tool Usage Awareness~\cite{huang2023metatool,huang2024planning}{, }Pass Rate~\cite{qin2023toolllm}{, } Accuracy~\cite{song2023restgpt,chen2023t}{.}\\
                            Tool Selection:~Recall~\cite{zhu2004recall}{, }NDCG~\cite{jarvelin2002cumulated}{, }COMP~\cite{qu2024towards}{.}\\Tool Calling:~Consistent with stipulations~\cite{chen2023t,ye2024tooleyes,huang2024planning}{.}\\Response Generation:~BLEU~\cite{papineni-etal-2002-bleu}{, }ROUGE-L~\cite{lin2004rouge}{, }Exact Match~\cite{blackwell2009cem}{.}, leaf, text width=40.5em
                        ]
                    ]
                ]
                [
                    Challenges \\and Future\\ Directions~(\S \ref{sec:challenge})
                    [
                        (1)~High Latency in Tool Learning~(\S\ref{subsec:high latency}){, }
                        (2)~Rigorous and Comprehensive Evaluation~(\S\ref{subsec:Rigorous and Comprehensive Evaluation}){, }
                        (3)~Comprehensive and Accessible Tools~(\S\ref{subsec:Comprehensive and Accessible Tools}){, }\\(4)~Safe and Robust Tool Learning~(\S\ref{subsec:Safe and Robust Tool Learning}){, }
                        (5)~Unified Tool Learning Framework~(\S\ref{subsec:Unified Tool Learning Framework}){, }
                        (6)~Real-Word Benchmark for Tool Learning~(\S\ref{subsec:Real-Word Benchmark for Tool Learning}){, }\\
                        (7)~Tool Learning with Multi-Modal~(\S\ref{subsec:Tool Learning with Multi-Modal}){.}
                        , leaf, text width=48.4em
                    ]
                ]
            ]
        \end{forest}
    }
    \caption{The overall structure of this paper.}
    \label{fig:structure}
    \vspace{.2cm}
\end{figure*}
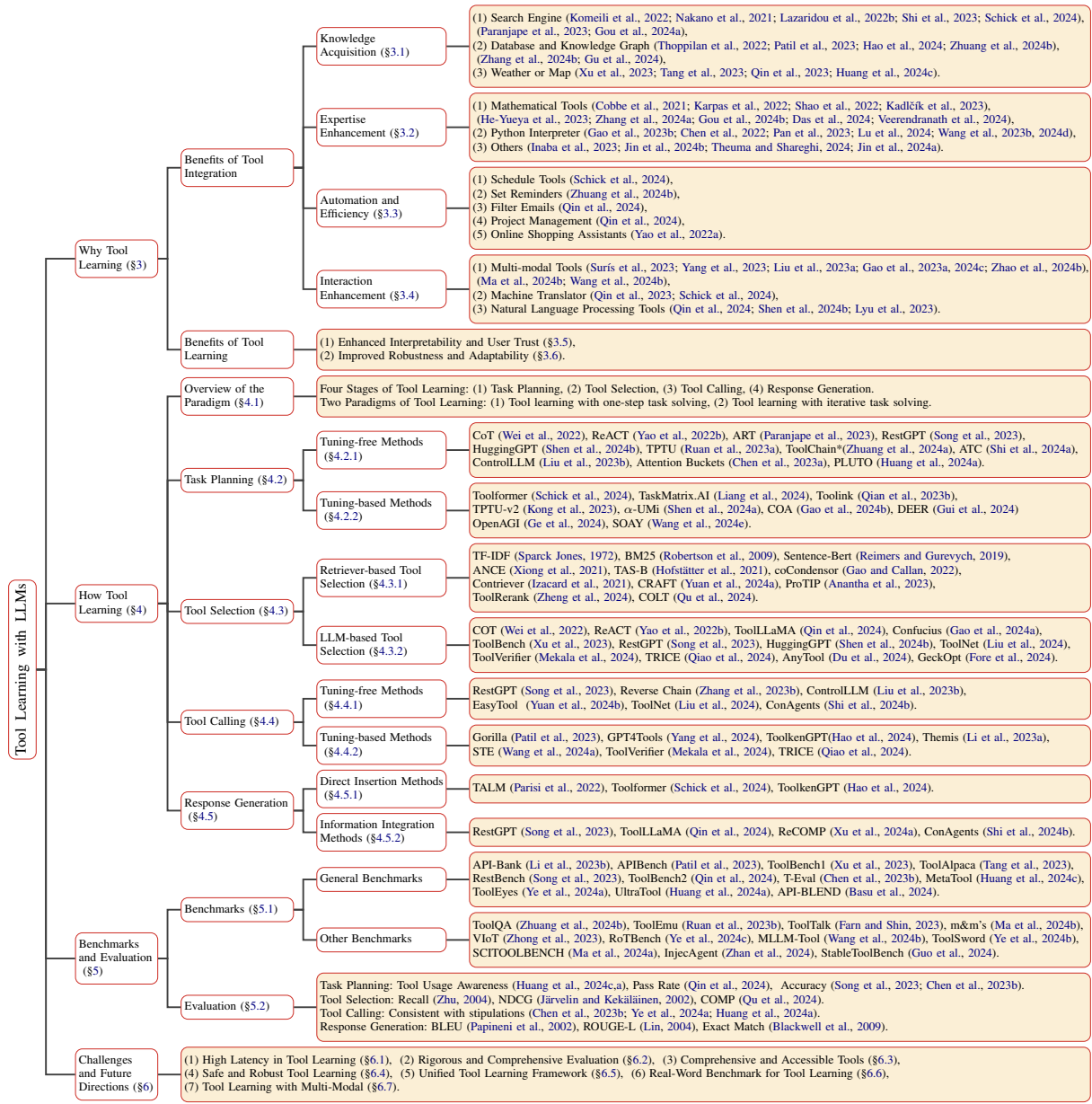

The remaining part of this paper~(as illustrated in Figure~\ref{fig:structure}) is organized as follows: We begin by introducing the foundational concepts and terminology related to tool learning~(\S\ref{sec:background}). 
Following this, we explore the significance of tool learning for LLMs from six specific aspects~(\S\ref{sec:why}).
We then systematically review the recent advancements in tool learning, focusing on four distinct stages of the tool learning workflow~(\S\ref{sec:how tool learning}). 
Subsequently, we provide a summary of the resources available for tool learning, including benchmarks and evaluation methods~(\S\ref{sec:benchmarks and evaluation}). 
Next, we discuss the current challenges in the field and outline open directions for future research~(\S\ref{sec:challenge}).
Lastly, we conclude the survey by summarizing the key findings~(\S\ref{sec:conclusion}).

\section{Background}
\label{sec:background}

In this section, we provide an overview of the key concepts and terminology related to tool learning, offering a clearer understanding of the fundamental aspects of this field.

\paratitle{What is a Tool?} 
The definition of a tool is notably broad within the context of augmented LLMs. Mialon et al. (2023)~\cite{mialon2023augmented} articulate a tool as ``\textit{the external employment of an unattached or manipulable attached environmental object to alter more efficiently the form, position, or condition of another object.}''
On the other hand, Wang et al. (2024)~\cite{wang2024tools} define a tool as ``\textit{An LM-used tool is a function interface to a computer program that runs externally to the LM, where the LM generates the function calls and input arguments in order to use the tool.}''
Similarly, it is our contention that any method enhancing LLMs through external means qualifies as a tool. 
Notably, retrieval-augmented generation~(RAG) represents a specific instance of tool learning, wherein the search engine is employed as a tool for LLMs.
Meanwhile, the definition of ``tool'' often remains vague and inconsistent across different papers.
For example, some studies distinctly define tools and APIs, positing that a tool comprises an aggregation of multiple APIs~\cite{patil2023gorilla,xu2023tool,qin2023toolllm}. 
Conversely, other studies treat each API as an independent tool~\cite{anantha2023protip,li2023api,tang2023toolalpaca}. 
In this survey, adhering to the definitions of tools established earlier in the text, we consider each API as an individual tool.

\paratitle{What is Tool Learning?} 
Tool learning refers to the process that ``\textit{aims to unleash the power of LLMs to effectively interact with various tools to accomplish complex tasks}''~\cite{qin2023toolllm}. 
This paradigm significantly improves the ability of LLMs to solve complex problems.
For example, when ChatGPT receives a user query, it evaluates the necessity of calling a specific tool. 
If a tool is required, ChatGPT will transparently outline the problem-solving process using the tool, explaining the rationale behind its responses, thereby ensuring the user receives a well-informed answer. 
Moreover, in instances where the initial solution fails, ChatGPT will reassess its tool selection and employ an alternative to generate a new response.

\section{Why Tool Learning?}
\label{sec:why}
In this section, we will delineate the multifaceted importance of tool learning for LLMs from two principal perspectives: the benefits of tool integration and the benefits of the tool learning paradigm itself. On the one hand, tool integration into LLMs enhances capabilities across several domains, namely knowledge acquisition, expertise enhancement, automation and efficiency, and interaction enhancement. On the other hand, the adoption of the tool learning paradigm bolsters the robustness of responses and transparency of generation processes, thereby enhancing interpretability and user trust, as well as improving system robustness and adaptability. Subsequent subsections will elaborate on these six aspects in detail, outlining why tool learning is important for LLMs.

\subsection{Knowledge Acquisition}
\label{subsec:knowledge Acquisition}
Although LLMs have showcased their immense capabilities across various fields~\cite{ouyang2022training}, their abilities are still bounded by the extent of knowledge learned during pre-training~\cite{mallen2022not}.
This embedded knowledge is finite and lacks the ability to acquire updated information. 
Additionally, the effectiveness of LLMs is further compromised by prompts from users, which may not always be meticulously crafted. 
Consequently, LLMs are prone to generating contents that seem superficially plausible but may contain factual inaccuracies, which is known as hallucination.
A promising approach to mitigate these limitations involves augmenting LLMs with the capability to access external tools, which allows LLMs to acquire and integrate external knowledge dynamically.
For example, the employment of search engine tool can enable LLMs to access contemporary information~\cite{komeili2021internet,nakano2021webgpt,zhang2023toolcoder,shi2023replug,schick2024toolformer,paranjape2023art,gou2023critic}, while the integration of database tool allows LLMs to access structured databases to retrieve specific information or execute complex queries, thus expanding their knowledge base~\cite{thoppilan2022lamda,patil2023gorilla,hao2024toolkengpt,zhuang2024toolqa,zhang2024syntax,gu2024middleware}. 
Additionally, connections to weather tools allow for real-time updates on weather conditions, forecasts, and historical data \cite{xu2023tool,tang2023toolalpaca,huang2023metatool}, and interfacing with mapping tools enables LLMs to get and provide geographical data, aiding in navigation and location-based queries \cite{qin2023tool}. 
Through these enhancements, LLMs can surpass traditional limitations, offering more accurate and contextually relevant outputs.

\subsection{Expertise Enhancement}
\label{subsec:expertise}
Given the fact that LLMs are trained on datasets comprising general knowledge, they often exhibit deficiencies in specialized domains.
While LLMs demonstrate robust problem-solving capabilities for basic mathematical problems, excelling in operations such as addition, subtraction, and exhibiting reasonable proficiency in multiplication tasks, their abilities significantly decline when confronted with division, exponentiation, logarithms, trigonometric functions, and other more complex composite functions~\cite{dao2023investigating,wei2023cmath}.
This limitation extends to tasks involving code generation~\cite{chen2021evaluating,austin2021program} and chemistry and physics problems~\cite{inaba-etal-2023-multitool,m2024augmenting}, \etc., further underscoring the gap in their expertise in more specialized areas.
Consequently, it is feasible to employ specific tools to augment the domain-specific expertise of LLMs~\cite{he2023solving,kadlvcik2023calc,jin2024genegpt,singh2024evaluating}.
For example, LLMs can use online calculators or mathematical tools to perform complex calculations, solve equations, or analyze statistical data~\cite{cobbe2021training,karpas2022mrkl,shao2022chaining,kadlvcik2023calc,he2023solving,zhang2024evaluating,gou2023tora,das2024mathsensei,veerendranath2024calccmu,bulusu2024mathviz}.
Additionally, the integration of external programming resources such as Python compilers and interpreters allows LLMs to receive code execution feedback, which is essential for refining code to align with user requirements and to optimize the code generation~\cite{gao2023pal,chen2022program,pan-etal-2023-fact,lu2024chameleon,wang2023leti,wang2023mint,wu2024structureaware,zhang2024codeagent}. 
Moreover, LLMs can also leverage tools in fields such as chemistry~\cite{inaba-etal-2023-multitool,m2024augmenting,ramos2024review}, biology~\cite{jin2024genegpt}, economics~\cite{theuma2024equipping,gao2024simulating,zhang2024multimodal}, medicine~\cite{jin2024agentmd,li2024mmedagent}, and recommendation systems~\cite{zhao2024let} to enhance their domain-specific expertise.
This approach not only mitigates the expertise gap in LLMs but also enhances their utility in specialized applications by providing domain-specific knowledge.

\subsection{Automation and Efficiency}
\label{subsec:action}
LLMs are fundamentally language processors that lack the capability to execute external actions independently, such as reserving conference rooms or booking flight tickets~\cite{wang2024tools}.
The integration of LLMs with external tools facilitates the execution of such tasks by simply populating tool interfaces with the necessary parameters. 
For example, LLMs can employ task automation tools to automate repetitive tasks such as scheduling~\cite{schick2024toolformer}, setting reminders~\cite{zhuang2024toolqa}, and filtering emails~\cite{qin2023toolllm}, thereby enhancing their practicality for user assistance. 
Moreover, by interfacing with project management and workflow tools, LLMs can aid users in managing tasks, monitoring progress, and optimizing work processes~\cite{qin2023toolllm}.
In addition, the integration with online shopping assistants not only simplifies the shopping process~\cite{yao2022webshop} but also enhances processing efficiency and user experience. 
Furthermore, employing data table processing tools enables LLMs to perform data analysis and visualization directly~\cite{qin2023tool}, thereby simplifying the data manipulation process of users.

\subsection{Interaction Enhancement}
\label{subsec:multi-modal}
Due to the diverse and multifaceted nature of user queries in the real-world, which may encompass multiple languages and modalities, LLMs often face challenges in consistently understanding different types of input. 
This variability can lead to ambiguities in discerning the actual user intent~\cite{wang2024tool}. 
The deployment of specialized tools can significantly enhance the perceptual capabilities of LLMs. 
For example, LLMs can utilize multi-modal tools, such as speech recognition and image analysis, to better understand and respond to a broader spectrum of user inputs~\cite{suris2023vipergpt,yang2023mm,liu2023internchat,gao2023assistgpt,gao2023clova,zhao2024diffagent,ma2024m,wang2024tool}.
Moreover, by interfacing with machine translator tools, LLMs have the capability to convert languages in which they are less proficient into languages they comprehend more effectively~\cite{schick2024toolformer,qin2023tool}.
Additionally, the integration of advanced natural language processing tools can augment the linguistic understanding of LLMs, thereby optimizing dialogue management and intent recognition~\cite{qin2023toolllm,shen2024hugginggpt,lyu2023gitagent}. 
Such advancements may include platforms that utilize contextual understanding models to elevate the performance of chatbot systems. 
Ultimately, improving perceptual input and sensory perception is crucial for the progression of LLMs capabilities in managing intricate user interactions.

\begin{figure*}[t]
\centering
	\includegraphics[width=\linewidth]{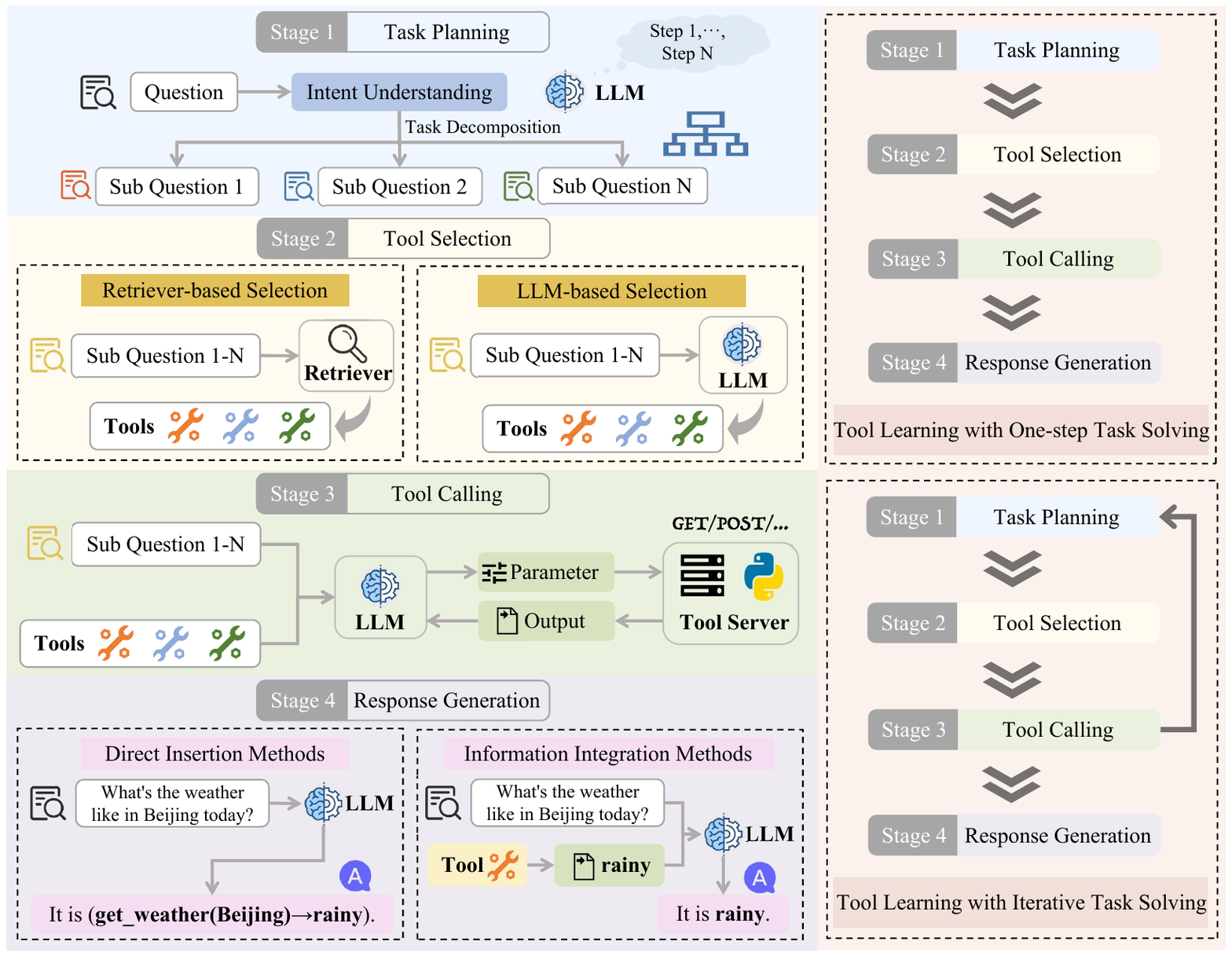}
        \caption{The overall workflow for tool learning with large language models. The left part illustrates the four stages of tool learning: task planning, tool selection, tool calling, and response generation. The right part shows two paradigms of tool learning: Tool Learning with One-step Task Solving and Tool Learning with Iterative Task Solving.}
        \label{fig: workflow}
\end{figure*}

\subsection{Enhanced Interpretability and User Trust}
\label{subsec:interpretability}
A significant concern with current LLMs is their opaque, ``black-box'' nature, which does not reveal the decision-making process to users~\cite{linardatos2020explainable,zhao2024explainability}, thereby severely lacking in interpretability. 
This opacity often leads to skepticism about the reliability of the response provided by LLMs and makes it challenging to ascertain their correctness~\cite{weidinger2021ethical}. 
Moreover, interpretability is particularly crucial in high-stakes domains such as aviation, healthcare and finance~\cite{qin2023tool,theuma2024equipping}, where accuracy is imperative. 
Therefore, understanding and explaining LLMs is crucial for elucidating their behaviors~\cite{zhao2024explainability}.
Some studies have enhanced the accuracy and interpretability of LLMs by enabling them to generate text with citations~\cite{gao-etal-2023-enabling,sun2023towards}. 
In contrast, through the utilization of tool learning, LLMs can exhibit each step of their decision-making process, thereby making their operations more transparent~\cite{qin2023tool}. 
Even in cases of erroneous outputs, such transparency allows users to quickly identify and understand the source of errors,
which facilitates a better understanding and trust in the decisions of LLMs, thus enhancing effective human-machine collaboration.

\subsection{Improved Robustness and Adaptability}
\label{subsec:robustness}

Existing research indicates that LLMs are highly sensitive to user inputs within prompts~\cite{wallace-etal-2019-universal,jin2020bert,wu2024new}. 
Merely minor modifications to these inputs can elicit substantial changes in the responses, highlighting a lack of robustness in LLMs. 
In the real world, different users have varying interests and ways of asking questions, leading to a diverse array of prompts.
The integration of specialized tools has been proposed as a strategy to reduce reliance on the statistical patterns in the training data~\cite{qin2023tool,shen2024hugginggpt,schick2024toolformer,qin2023toolllm,hao2024toolkengpt}. 
Though the input format from the user is different, the input and output of the tool are the same.
This enhancement increases the resistance of LLMs to input perturbations and their adaptability to new environments. 
Thus, such integration not only stabilizes the models in uncertain conditions but also reduces the risks associated with input errors.

\section{How Tool Learning?}
\label{sec:how tool learning}
In this section, we will first introduce the overall paradigm of tool learning, which includes four distinct stages and two typical paradigms.
Following this framework, we provide a detailed review of each stage within the tool learning workflow, along with the latest advancements associated with each stage. It's important to note that many works involve multiple stages of tool learning, but we only discuss its core stages here. For each stage, we also present a practical, real-world example utilizing GPT-4 for tool learning to address a specific problem, which are designed to help newcomers better understand what each stage involves and how it is implemented.

\subsection{Overall Paradigm of Tool Learning}
\label{subsec:overview}
In this section, we will introduce the entire process of tool learning, including four stages and two paradigms involved in the utilization of tool-augmented LLMs.

\paratitle{Four Stages of Tool Learning.}
As illustrated in the left part of Figure~\ref{fig: workflow}, the typical process of tool learning comprises four stages: task planning, tool selection, tool calling, and response generation, which is adopted in numerous works related to tools~\cite{song2023restgpt,shen2024hugginggpt,ruan2023tptu}. 
This process outlines the user interaction pipeline with tool-augmented LLMs: 
given a user question, the preliminary stage involves the LLMs analyzing the requests of users to understand their intent and decompose it into potential solvable sub-questions.
Subsequently, the appropriate tools are selected to tackle these sub-questions.
This tool selection process is categorized into two types based on whether a retriever is used: retriever-based tool selection and LLM-based tool selection.
Recently, there has been an increasing focus on initially using a retriever to filter out the top-k suitable tools~\cite{qin2023toolllm,Gao2023ConfuciusIT,anantha2023protip}. 
This necessity stems from the fact real-world systems usually have a vast number of tools, rendering it impractical to incorporate the descriptions of all tools as input for LLMs due to the constraints related to length and latency~\cite{qu2024towards}. 
Subsequently, the user query along with the selected tools are furnished to the LLMs, enabling it to select the optimal tool and configure the necessary parameters for tool calling.
This necessitates that the LLMs possess a keen awareness of using tools and be able to correctly select the tools needed.
Moreover, it is imperative for the LLMs to extract the correct tool parameters from the user query, a process that demands not only the accuracy of the parameter content but also adherence to the specific format requirements. 
Following the invocation of the tool, the LLMs utilizes the results returned by the tool to craft a superior response for the user.

\paratitle{Two Paradigms of Tool Learning.}
As illustrated in the right part of Figure~\ref{fig: workflow}, the paradigms for employing tool learning can be categorized into two types: tool learning with one-step task solving and tool learning with iterative task solving. These are also referred to as planning without feedback and planning with feedback in Wang et al. (2024)~\cite{wang2024survey}, and decomposition-first and interleaved decomposition in Huang et al. (2024)~\cite{huang2024understanding}. 
In earlier studies on tool learning~\cite{schick2024toolformer,shen2024hugginggpt,lu2024chameleon}, the primary paradigm is tool learning with one-step task solving: upon receiving a user question, LLMs would analyze the requests of user to understand the user intent and immediately plan all the sub-tasks needed to solve the problem.
The LLMs would then directly generate a response based on the results returned by the selected tools without considering the possibility of errors during the process or altering the plan based on tool feedback.
Subsequent researches introduce a new paradigm known as tool learning with iterative task solving~\cite{li2023api,song2023restgpt,qin2023toolllm,Gao2023ConfuciusIT}. 
This approach does not commit to a complete task plan upfront. 
Instead, it allows for iterative interactions with the tools, adjusting the sub-tasks progressively based on tool feedback. 
This enables LLMs to address the problem step-by-step, refining its plan continuously in response to the results returned by tools, thereby improving problem-solving capabilities. This shift from a one-step to an iterative approach marks a significant advancement in the field of tool learning.

\subsection{Task Planning}
\label{subsec:planning}
Generally, in real-world scenarios, user queries often embody complex intent. 
Thus, it is essential to first engage in task planning to conduct a comprehensive analysis of the user intent. 
This stage involves the decomposition of a user question into multiple sub-questions as required to achieve the specified objective. 
Additionally, given the dependency relationships inherent among some tasks, the planner is also tasked with delineating the dependencies and execution sequence of the decomposed tasks, thereby facilitating the establishment of interconnections between the sub-questions.
For example, a real example for task planning with GPT-4 is shown as follows:
\begin{tcolorbox}[title=An Example for Task Planning with GPT-4]
\small
\setstretch{1.2}
\textbf{Instruction Prompt:} You are currently in the task planning stage. You are given a user query requiring multi-step actions and reasoning. You will break down the user's query into sub-questions, and you only need to output these sub-questions after the breakdown. Ensure that the original problem is comprehensively covered through the minimal number of sub-questions. \\
\textbf{User Question:} \emph{I would like to know the value of 5 ounces of gold plus 1 million AMZN stocks in CNY.} \\ 
\textbf{Output:} \color{red}{\emph{1.What is the current price of gold per ounce in USD?
2.What is the current stock price of Amazon (AMZN) per share in USD?
3.What is the current exchange rate between USD and CNY (Chinese Yuan)?}}
\end{tcolorbox}

Tool learning with iterative task solving paradigm is not convenient for staged demonstrations, so this example uses the tool learning with one-step task solving paradigm. 
From this example, we can see that GPT-4 breaks down the user's question into three sub-questions. 
Since internal knowledge of GPT-4 does not contain the latest information, it is necessary to obtain the current prices of gold and Amazon (AMZN) stock, as well as the latest exchange rate between USD and CNY. 
By breaking down the query in this manner, GPT-4 can address complex questions step by step, demonstrating its logical analysis capabilities and ability to handle multi-step tasks.
Next, we will introduce the latest developments in two categories: tuning-free methods and tuning-based methods.

\subsubsection{Tuning-free Methods}
\label{subsubsec:planning_free}
Existing studies~\cite{paranjape2023art,zhang2023graph,li2024stride} demonstrate that the innate abilities of LLMs enable effective planning through methods such as few-shot or even zero-shot prompting. 
For example, some studies~\cite{huang2022language,chern2023factool,xu2023search,kim2024language} leverage prompts to decompose complex tasks into simpler sub-tasks, facilitating a structured plan of action. 
ART~\cite{paranjape2023art} constructs a task library, from which it retrieves examples as few-shot prompts when encountering real-world tasks.
RestGPT~\cite{song2023restgpt} introduces a Coarse-to-Fine Online Planning approach, an iterative task planning methodology that enables LLMs to progressively refine the process of task decomposition. 
HuggingGPT~\cite{shen2024hugginggpt} leverages a sophisticated prompt design framework, which integrates specification-based instructions with demonstration based parsing methods. 
ToolChain*~\cite{zhuang2023toolchain} employs a planning mechanism by constructing the entire action space as a decision tree, where each node within the tree represents a potential API function call.
TPTU~\cite{ruan2023tptu} introduces a structured framework specifically designed for LLM-based AI agents, incorporating two distinct types of agents: the One-step agent and the sequential agent. 
Attention Buckets~\cite{chen2023fortify} operates in parallel with unique RoPE angles, forming distinct waveforms that compensate for each other, reducing the risk of LLMs missing critical information.
ControlLLM~\cite{liu2023controlllm} introduces a paradigm known as Thoughts-on-Graph (ToG), which leverages Depth-First Search (DFS) on a pre-constructed tool graph to identify solutions. 
PLUTO~\cite{huang2024planedit} uses an autoregressive planning approach to iteratively improve performance by generating hypotheses, performing cluster analysis, and refining sub-queries until the initial query is satisfied.
ATC~\cite{shi2024chain} enables LLMs to independently learn and master new tools by using a chain of tools and a black-box probing method to identify and record tool usage.
Tool-Planner~\cite{liu2024tool} organizes tools into toolkits based on API functions with similar functionality, enabling LLMs to plan across different toolkits.
SGC~\cite{wu2024can} enhances task planning by integrating GNNs with LLMs, enabling more efficient and accurate sub-task selection within task graphs.
Sum2Act~\cite{liu2024summary} guides LLMs to solve complex tasks by summarizing progress at each step and adjusting actions based on task state and errors.
BTP~\cite{zheng2024budget} creates an optimal plan for tool usage under budget constraints, allowing LLMs to efficiently manage costs while solving user queries.
DRAFT~\cite{qu2024from} proposes a novel framework aimed at dynamically adjusting and optimizing tool documentation based on the interaction feedback between LLMs and external tools, thereby improving LLMs' comprehension and tool-using capabilities.

\subsubsection{Tuning-based Methods}
\label{subsubsec:planning_tune}
Though LLMs demonstrate impressive performance in zero-shot or few-shot settings, they remain less effective compared to models that have been fine-tuned~\cite{erbacher2024navigating}.
Toolformer~\cite{schick2024toolformer} employs API calls that actually assist the model in predicting future tokens to fine-tune GPT-J, which enhances the awareness and capability of LLMs to utilize tools effectively.
TaskMatrix.AI~\cite{liang2024taskmatrix} leverages Reinforcement Learning from Human Feedback (RLHF) to utilize the knowledge and insights gained through human feedback, thereby enhancing the foundation model.
Toolink~\cite{qian2023toolink} innovates by decomposing the target task into a toolkit for problem-solving, then employing a model to utilize these tools to answer queries via a chain-of-solving (CoS) approach. 
TPTU-v2~\cite{kong2023tptuv2} develops an LLM finetuner to fine-tune a base LLM using a meticulously curated dataset, so that the finetuned LLM can be more capable of task planning and API calls, especially for domain-specific tasks.
$\alpha$-UMi~\cite{shen2024small} presents a novel two-phase training paradigm where a foundational large language model is first extensively fine-tuned and then replicated as a planner for further fine-tuning on planning tasks.
COA~\cite{gao2024efficient} trains LLMs to first decode reasoning chains with abstract placeholders, and then call domain tools to reify each reasoning chain by filling in specific knowledge.
DEER~\cite{gui2024look} stimulates decision-making awareness in LLMs across various scenarios by automatically generating tool usage examples with multiple decision branches, and enhances the generalization ability of LLMs towards unseen tools through proposing novel tool sampling strategies.
SOAY~\cite{wang2024solution} first lets the LLM generate a feasible API calling plan, i.e. solution, based on complex user inputs, and then allows the LLM to generate executable API calling code based on the generated solution.
TP-LLaMA~\cite{chen2024advancing} introduces a method to create preference data from thought trees by using previously ignored failed explorations, creating a dataset for DPO to update the LLM's policy.
APIGen~\cite{liu2024apigen} creates an automated pipeline for generating diverse, high-quality, and verifiable function-calling datasets, significantly enhancing the performance of models fine-tuned with this data.

\paratitle{Remark.}
In summary, task planning, as the initial stage of tool learning, is crucial for solving the entire problem. 
Although there are many methods currently available to enhance the task planning capabilities of LLMs, generating a perfect plan directly when facing complex issues remains challenging. 
Furthermore, tool learning is a process involving interaction between LLMs and tools. 
How to better utilize feedback from tools to improve planning is still a question worthy of investigation.

\subsection{Tool Selection}
\label{subsec:selection}

After the task planning phase, LLMs have already decomposed the user question into multiple sub-questions. 
In order to better address these sub-questions, it is necessary to select appropriate tools. 
The tool selection process involves choosing through a retriever or directly allowing LLMs to pick from a provided list of tools.
When there are too many tools, a tool retriever is typically used to identify the top-$K$ relevant tools to offer to the LLMs, a process known as retriever-based tool selection. 
If the quantity of tools is limited or upon receiving the tools retrieved during the tool retrieval phase, the LLMs need to select the appropriate tools based on the tool descriptions and the sub-question, which is known as LLM-based tool selection.
For example, an example for tool selection with GPT-4 is shown as follows:
\begin{tcolorbox}[title=An Example for Tool Selection with GPT-4] 
\small
\setstretch{1.2}
\textbf{Instruction Prompt:} You are currently in the tool selection stage. You are given candidate tools that can be potentially used to solve the sub-question. Among candidate tools, select a list of relevant tools that would help solve the sub-question.\\
\textbf{Sub-question 1:} \emph{What is the current price of gold per ounce in USD?} \\
\textbf{Candidate Tools:} \emph{1.Metals Prices Rates API: The latest API endpoint will return real-time exchange rate data updated every 60 seconds.
2.Medium: Get official news from Medium.
3.Cryptocurrency Markets: Recently published cryptocurrencies videos.}\\
\textbf{Output:} \color{red}{\emph{ 1.Metals Prices Rates API: The latest API endpoint will return real-time exchange rate data updated every 60 seconds.\\}}
\color{black}\textbf{Sub-question 2:} {$\cdots$} \\ 
\color{black}{\quad \quad \quad \quad \quad \quad \quad \quad \quad \quad \quad \quad $\vdots$} \\
\color{black}\textbf{Output:} {$\cdots$}
\end{tcolorbox}

From this example, we can see that for the sub-question about obtaining the price of gold, GPT-4 can correctly select the necessary tools. 
Specifically, when faced with multiple candidate tools, GPT-4 can analyze the features of each tool and choose the one most suitable for answering the question.
In this example, GPT-4 selects the Metals Prices Rates API because it provides real-time updated information on gold prices. This demonstrates accuracy and effectiveness of GPT-4 in tool selection.
Next, we will introduce the latest developments in two categories: retriever-based tool selection and LLM-based tool selection.

\subsubsection{Retriever-based Tool Selection}
\label{subsec:retriever}

Real-world systems often incorporate a wide array of tools, making it impractical to input descriptions of all tools into LLMs due to length limitations and latency constraints.
Therefore, to fully exploit the potential of tool-augmented LLMs, it is crucial to develop an efficient tool retrieval system. 
This system aims to bridge the gap between the broad capabilities of LLMs and the practical limitations of input size by efficiently selecting the top-$K$ most suitable tools for a given query from a vast tool set.
State-of-the-art retrieval methods can be categorized into two types: term-based and semantic-based. 

\paratitle{Term-based Methods.}
Term-based methods~(\ie sparse retrieval) represent both documents and queries as high-dimensional sparse vectors based on terms, as exemplified by TF-IDF~\cite{sparck1972statistical} and BM25~\cite{robertson2009probabilistic}.
These methods employ exact term matching to achieve efficient alignment between queries and documents.
For example, Gorilla~\cite{patil2023gorilla} employs BM25 and GPT-Index to construct a retriever for implementing tool retrieval.

\paratitle{Semantic-based Methods.}
Conversely, semantic-based methods~(\ie dense retrieval) utilize neural networks to learn the semantic relationship between queries and tool descriptions~\cite{reimers2019sentence,xiong2020approximate,hofstatter2021efficiently,gao2021unsupervised,izacard2021unsupervised}, and then calculate the semantic similarity using methods such as cosine similarity.
Recently, there has been a burgeoning interest in the development and refinement of more efficient tool retrievers. 
Some studies~\cite{kong2023tptuv2,qin2023toolllm,Gao2023ConfuciusIT} train a Sentence-Bert model as the tool retriever, enabling the high-efficiency retrieval of relevant tools.
CRAFT~\cite{yuan2023craft} instructs LLMs to generate a fictitious tool description based on the given query and then employs this fabricated tool to conduct a search.
Anantha et al. (2023)~\cite{anantha2023protip} propose ProTIP based on the concept of task decomposition.
Xu et al. (2024a)~\cite{xu2024enhancing} propose a method that enhances tool retrieval by leveraging feedback from LLMs to progressively refine instructions and align retrieval with tool usage.
COLT~\cite{qu2024towards} proposes a novel tool retrieval approach using GNNs, identifying that a critical dimension often overlooked in conventional tool retrieval methodologies is the necessity to ensure the completeness of the tools retrieved.
In addition to the recall phase, Zheng et al. (2024)~\cite{zheng2024toolrerank} also take into account the re-ranking stage of tool retrieval. 
They consider the differences between seen and unseen tools, as well as the hierarchical structure of the tool library. 
Building on these considerations, they propose an adaptive and hierarchy-aware Re-ranking method, ToolRerank.
Meanwhile, we can also directly employ off-the-shelf embeddings~\cite{bge_embedding,geminiteam2024gemini} to get the representations of user queries and tool descriptions.
In conclusion, constructing an efficient tool retriever is of paramount importance.

\paratitle{Remark.}
Although traditional information retrieval methods are suitable for tool retrieval scenarios, they still have issues such as focusing solely on semantic similarity and ignoring the hierarchical structure of the tools, \etc. 
Future work should consider the unique needs and characteristics specific to tool retrieval scenarios in order to build a more effective tool retriever.

\subsubsection{LLM-based Tool Selection}
\label{subsec:llm selection}
In instances where the quantity of tool libraries is limited or upon receiving the tools retrieved from the tool retrieval phase, it is feasible to incorporate the descriptions and parameter lists of these tools into the input context along with the user query provided to LLMs.
Subsequently, LLMs are tasked with selecting the appropriate tools from the available tool list based on the user query. 
Given that the resolution of queries is occasionally sensitive to the order in which tools are invoked, there is a necessity for serial tool calling, where the output of one tool may serve as the input parameter for another. 
Consequently, this demands a high degree of reasoning capability from the LLMs. 
It must adeptly select the correct tools based on the information currently at its disposal and the information that needs to be acquired.
Existing methods can be similarly categorized into tuning-free and tuning-based approaches.

\paratitle{Tuning-free Methods.}
Tuning-free methods capitalize on the in context learning ability of LLMs through strategic prompting~\cite{song2023restgpt,shen2024hugginggpt}. 
For instance, Wei et al. (2022)~\cite{wei2022chain} introduce the concept of chain of thought (COT), effectively incorporating the directive ``let's think step by step'' into the prompt structure. 
Further advancing this discourse, Yao et al. (2022)~\cite{yao2022react} propose ReACT, a framework that integrates reasoning with action, thus enabling LLMs to not only justify actions but also to refine their reasoning processes based on feedback from the environment~(\eg the output of tools). 
This development marks a significant step forward in enhancing the adaptability and decision-making capabilities of LLMs by fostering a more dynamic interaction between reasoning and action.
Building upon these insights, Qin et al. (2024)~\cite{qin2023toolllm} propose DFSDT method, which addresses the issue of error propagation by incorporating a depth-first search strategy to improve decision-making accuracy.
ChatCoT~\cite{chen-etal-2023-chatcot} integrates tool use into multi-turn reasoning, allowing LLMs to seamlessly combine conversation-based reasoning with tool manipulation.
ToolNet~\cite{liu2024toolnet} organizes tools into a directed graph, enabling LLMs to navigate from an initial tool node, iteratively selecting tools until the task is completed.
AnyTool~\cite{du2024anytool} designs a more efficient tool retriever by leveraging the hierarchical structure of tools.
GeckOpt~\cite{fore2024geckopt} narrows down tool selection by adding intent-driven gating.

\paratitle{Tuning-based Methods.}
Tuning-based methods directly fine-tune the parameters of LLMs on the tool learning dataset to master tool usage.
Toolbench~\cite{xu2023tool} analyzes the challenges faced by open-source LLMs during the tool learning process, suggesting that fine-tuning, along with utilizing demonstration retrieval and system prompts, can significantly enhance the effectiveness of LLMs in tool learning.
TRICE~\cite{qiao2023making} proposes a two-stage framework, which initially employs behavior cloning for instruct-tuning of the LLMs to imitate the behavior of tool usage, followed by further reinforcing the model through RLEF by utilizing the tool execution feedback.
ToolLLaMA~\cite{qin2023toolllm} employs the instruction solution pairs derived from DFSDT method to fine-tune the LLaMA $7$B model.
Confucius~\cite{Gao2023ConfuciusIT} acknowledges the diversity in tool complexity and proposes a novel tool learning framework. 
ToolVerifier~\cite{mekala2024toolverifier} introduces a self-verification method which distinguishes between close candidates by self-asking contrastive questions during tool selection.

\paratitle{Remark.}
By comparing the aforementioned methods, we can find that the tuning-based method improves the capability of LLMs in tool selection by modifying model parameters. 
This approach can integrate extensive knowledge about tools, but it is only applicable to open-source LLMs and incurs substantial computational resource consumption. 
Conversely, the tuning-free method enhances the capability of LLMs in tool selection using precise prompting strategies or by modifying existing mechanisms, and it is compatible with all LLMs.
However, since the possibilities for designing prompts are limitless, finding the ideal way to create the perfect prompt is still a major challenge.

In real-world applications, the sheer volume of available tools, coupled with constraints such as limited context length and high latency, makes it impractical to present all options to LLMs simultaneously. To address this challenge, traditional retrieval methods are typically employed first to filter and narrow down the pool of potential tools. This initial step is crucial for managing the complexity and ensuring that the subsequent LLM-based selection is both efficient and effective. After this retrieval process, the LLM can then refine the selection, making the final choice that aligns with its reasoning and preferences. This two-step approach highlights the complementary roles of both traditional retrieval methods and LLM-based techniques, demonstrating their collective importance in optimizing the tool selection process for real-world scenarios.

\subsection{Tool Calling}
\label{subsec:tool calling}
In the tool calling stage, LLMs need to extract the required parameters from the user query in accordance with the specifications outlined in the tool description and request data from tool servers.
This process mandates that the LLMs not only correctly extract the parameters' content and format but also adhere strictly to the prescribed output format to prevent the generation of superfluous sentences.
For example, an example for tool calling with GPT-4 is shown as follows:
\begin{tcolorbox}[title=An Example for Tool Calling with GPT-4]
\small
\setstretch{1.2}
\textbf{Instruction Prompt:} You are currently in the tool calling stage. You are given selected tools that can be potentially used to solve the sub-question. Your goal is to extract the required parameters needed to call the tool from the sub-question based on the tool descriptions. Output in the following format: \{parameter name: parameter, $\cdots$, parameter name: parameter\}\\
\textbf{Sub-question 1:} \emph{What is the current price of gold per ounce in USD?} \\
\textbf{Selected Tools:} \emph{Tool Name: \{Metals Prices Rates API\}. 
Tool description: \{The latest API endpoint will return real-time exchange rate data updated every 60 seconds.\} 
Required params:\{
[{name: symbols, type: STRING, description: Enter a list of comma-separated currency codes or metal codes to limit output codes.}, {name: base, type: STRING, description: Enter the three-letter currency code or metal code of your preferred base currency.}]
\}}\\
\textbf{Output:} \color{red}{\emph{\{symbols: ``XAU'', base: ``USD''\}\\}}
\color{black}\textbf{Sub-question 2:} {$\cdots$} \\ 
\color{black}{\quad \quad \quad \quad \quad \quad \quad \quad \quad \quad \quad \quad $\vdots$} \\
\color{black}\textbf{Output:} {$\cdots$}
\end{tcolorbox}

From this example, we can see that GPT-4 can extract the necessary parameters for calling a tool based on the provided user question and the selected tool's documentation. Specifically, GPT-4 can parse the critical information in the tool description and accurately identify which parameters need to be provided. 
Next, we will introduce the latest developments in the same way as the previous two stages, dividing them into tuning-free methods and tuning-based methods.

\subsubsection{Tuning-free Methods}
\label{subsubsec:calling_free}
Tuning-free methods predominantly leverage the few-shot approach to provide demonstrations for parameter extraction or rule-based methods, thereby enhancing the capability of LLMs to identify parameters~\cite{hsieh2023tool,song2023restgpt,liu2023controlllm,liu2024toolnet}. 
Reverse Chain~\cite{zhang2023reverse} utilizes reverse thinking by first selecting a final tool for a task and then having the LLMs populate the necessary parameters; if any are missing, an additional tool is chosen based on the description to complete them and accomplish the task.
EasyTool~\cite{yuan2024easytool} enhances the comprehension of LLMs regarding tool functions and parameter requirements by prompting ChatGPT to rewrite tool descriptions, making them more concise and incorporating guidelines for tool functionality directly within the descriptions.
Xu et al. (2024b)~\cite{xu2024concise} propose a method that compresses tool documentation into summary sequences while preserving key information, enabling efficient tool usage in LLMs with minimal performance loss.
ConAgents~\cite{shi2024learning} introduces a multi-agent collaborative framework, featuring a specialized execution agent tasked with parameter extraction and tool calling.

\subsubsection{Tuning-based Methods}
\label{subsubsec:call_tune}
Some studies enhance the tool calling capabilities of LLMs using tuning-based methods~\cite{patil2023gorilla,qiao2023making,mekala2024toolverifier}.
For example, GPT4Tools~\cite{yang2024gpt4tools} enhances the open-source LLMs by integrating tool usage capabilities through fine-tuning with LoRA optimization techniques, using a dataset of tool usage instructions generated by ChatGPT.
Toolkengpt~\cite{hao2024toolkengpt} uses special tokens called ``toolkens'' to seamlessly call tools, switching to a special mode upon predicting a toolken to generate required input parameters and integrate the output back into the generation process.
Themis~\cite{li2023tool} enhances the interpretability and scoring reliability of RMs by integrating tool usage and reasoning processes in an auto-regressive manner, dynamically determining which tools to call, how to pass parameters, and effectively incorporating the results into the reasoning process.
STE~\cite{wang2024llms} coordinates three key mechanisms in biological systems for the successful use of tools: trial and error, imagination, and memory, aiding LLMs in the accurate use of its trained tools.

Moreover, given the frequent occurrence of calling errors during the utilization of tools, such as incorrect formatting of input parameters, input parameters exceeding acceptable ranges of the tool, and tool server error, it is imperative to integrate error handling mechanisms. 
These mechanisms are designed to refine the action based on the error messages returned upon calling failure. 
This enables a more resilient and adaptive system, ensuring continuity and efficiency in tool learning even in the face of operational disruptions.

\paratitle{Remark.}
To sum up, although tuning-based methods can yield better results, they heavily rely on the tools seen in the training set, and perform poorly with new tools not included in the dataset.
Additionally, they face challenges such as catastrophic forgetting and a lack of robust generalization.
In contrast, tuning-free methods offer significant flexibility and do not require the construction of a specific dataset. 
So both tuning-based methods and tuning-free methods are crucial for enhancing the tool calling capabilities of LLMs.

\subsection{Response Generation}
\label{subsec:response generation}
Given the diversity and complexity of tool outputs, which can include various formats like text, numbers, code, videos, and images, it is often impractical to present these results directly to users.
Upon receiving outputs from tools, LLMs must synthesize information relevant to user queries and integrate their own knowledge to construct comprehensive responses.
For example, an example for response generation with GPT-4 is shown as follows:
\begin{tcolorbox}[title=An Example for Response Generation with GPT-4]
\small
\setstretch{1.2}
\textbf{Instruction Prompt:} You are currently in the response generation stage. You are given the output of selected tools that can be potentially used to solve the sub-task. Your goal is to generate a response to the user question based on the results from tools and your internal knowledge.\\
\textbf{User Question:} \emph{I would like to know the value of 5 ounces of gold plus 1 million AMZN stocks in CNY.} \\
\textbf{Return from Tool Calling:} \emph{The stock price of AMZN is $130$ USD, XAU USD is $1919.7605807522993$, and the exchange rate between USD and YNC is $7.24$.}\\
\textbf{Output:} \color{red}{\emph{
$\cdots$
Total value in CNY:
Total value = Value of gold in CNY + Value of stocks in CNY = $69,495.87$ + $941,200,000$ =$941,269,495.87$ CNY.
Therefore, the combined value of 5 ounces of gold and 1 million Amazon stocks in Chinese Yuan is approximately $941,269,495.87$ $\cdots$.}}
\end{tcolorbox}

From this example, we can see that GPT-4 can provide a better response by combining the results from tools and its internal knowledge. Specifically, GPT-4 uses real-time data provided by tools, such as the prices of gold and Amazon stock and the exchange rate, to calculate the final answer to the user's question. 
This demonstrates the ability of GPT-4 to integrate multiple information sources and perform complex calculations.
We categorize the latest advancements in this stage into two types: direct insertion methods and information integration methods.

\subsubsection{Direct Insertion Methods}
\label{subsubsec:response_direct}
The methods adopted in the early work involved directly inserting the output of tools into the generated response~\cite{parisi2022talm,schick2024toolformer,hao2024toolkengpt,wang2024tools}. 
For instance, if the user query is ``How is the weather today?'', LLMs produce a response like ``It's Weather()''~(as illustrated in Figure~\ref{fig: workflow}), which is subsequently replaced with the result returned by the tool~(\eg from ``It's Weather().'' to ``It's rainy.'').
However, given the outputs of tools are unpredictable, this method could potentially affect the user experience.

\begin{table*}[ht!]
\renewcommand\arraystretch{1.3}
\caption{A detailed list of different benchmarks and their specific configurations. Symbols \ding{172}, \ding{173}, \ding{174}, and \ding{175} represent the four stages in tool learning—task planning, tool selection, tool calling, and response generation, respectively.}
    \centering
    \resizebox{1\linewidth}{!}{
    \begin{tabular}{lccccccc}
    \hline 
        Benchmark & Focus & \# Tools & \# Instances & Tool Source & Multi-tool? & Executable? & Time \\ \hline
        \rowcolor{LightCyan} \multicolumn{8}{c}{\textbf{General Benchmarks}} \\ \hline
        API-Bank~\cite{li2023api} & \ding{172},  \ding{173}, \ding{174}, \ding{175} & $73$ & $314$ & Manual Creation & \cmark & \cmark & 2023-04 \\ 
        APIBench~\cite{patil2023gorilla} & \ding{173}, \ding{174} & $1,645$ & $16,450$ & Public Models & \xmark & \xmark & 2023-05 \\ 
        ToolBench1~\cite{xu2023tool} & \ding{173}, \ding{174} & $232$ & $2,746$ & Public APIs & \xmark & \cmark & 2023-05 \\ 
        ToolAlpaca~\cite{tang2023toolalpaca} & \ding{173}, \ding{174}, \ding{175} & $426$ & $3,938$ & Public APIs & \xmark & \xmark & 2023-06 \\ 
        RestBench~\cite{song2023restgpt} & \ding{172}, \ding{173}, \ding{174} & $94$ & $157$ & RESTful APIs & \cmark & \xmark & 2023-06 \\ 
        ToolBench2~\cite{qin2023toolllm} & \ding{172},  \ding{173}, \ding{174}, \ding{175} & $16,464$ & $126,486$ & Rapid API & \cmark & \cmark & 2023-07 \\ 
        MetaTool~\cite{huang2023metatool} & \ding{172},  \ding{173} & $199$ & $21,127$ & OpenAI Plugins & \cmark & \xmark & 2023-10 \\ 
        TaskBench~\cite{shen2023taskbench} & \ding{172}, \ding{173}, \ding{174} & $103$ & $28,271$ & Public APIs & \cmark & \cmark & 2023-11 \\ 
        T-Eval~\cite{chen2023t} & \ding{172}, \ding{173}, \ding{174} & $15$ & $533$ & Manual Creation & \cmark & \cmark & 2023-12 \\ 
        ToolEyes~\cite{ye2024tooleyes} & \ding{172}, \ding{173}, \ding{174}, \ding{175} & $568$ & $382$ & Manual Creation & \cmark & \cmark & 2024-01 \\ 
        UltraTool~\cite{huang2024planning} & \ding{172}, \ding{173}, \ding{174} & $2,032$ & $5,824$ & Manual Creation & \cmark & \xmark & 2024-01 \\ 
        API-BLEND~\cite{basu2024api} & \ding{173}, \ding{174} & - & $189,040$ & Exsiting Datasets & \cmark & \cmark & 2024-02 \\ 
        Seal-Tools~\cite{wu2024seal} & \ding{173}, \ding{174} & $4,076$ & $14,076$ & Manual Creation & \cmark & \xmark & 2024-05 \\ 
        ShortcutsBench	~\cite{shen2024shortcutsbench} & \ding{173}, \ding{174} & $1,414$ & $7,627$ &  Public APIs & \cmark & \cmark & 2024-07\\
        GTA~\cite{wang2024gta} & \ding{173}, \ding{174} \ding{175}& $14$ & $229$ &  Public APIs & \cmark & \cmark & 2024-07\\
        WTU-Eval~\cite{ning2024wtu} & \ding{172} & $4$ & $916$ &  BMTools & \cmark & \cmark & 2024-07\\
        AppWorld~\cite{trivedi2024appworld} &  \ding{172}, \ding{173}, \ding{174} & $457$ & $750$ &  FastAPI & \cmark & \cmark & 2024-07\\
        \hline
        \rowcolor{LightCyan} \multicolumn{8}{c}{\textbf{Other Benchmarks}} \\ \hline
        ToolQA~\cite{zhuang2024toolqa} & QA & $13$ & $1,530$ & Manual Creation & \xmark & \cmark & 2023-06 \\ 
        ToolEmu~\cite{ruan2023identifying} & Safety & $311$ & $144$ & Manual Creation & \xmark & \cmark & 2023-09 \\ 
        ToolTalk~\cite{farn2023tooltalk} & Conversation & $28$ & $78$ & Manual Creation & \xmark & \cmark & 2023-11 \\ 
        VIoT~\cite{zhong2023viotgpt} & VIoT & $11$ & $1,841$ & Public Models & \xmark & \cmark & 2023-12 \\ 
        RoTBench~\cite{ye2024rotbench} & Robustness & $568$ & $105$ & ToolEyes & \cmark & \cmark & 2024-01 \\ 
        MLLM-Tool~\cite{wang2024tool} & Multi-modal & $932$ & $11,642$ & Public Models & \cmark & \cmark & 2024-01 \\ 
        ToolSword~\cite{ye2024toolsword} & Safety & $100$ & $440$ & Manual Creation & \cmark & \cmark & 2024-02 \\ 
        SciToolBench~\cite{ma2024sciagent} & Sci-Reasoning & $2,446$ & $856$ & Manual Creation & \cmark & \cmark & 2024-02 \\ 
        InjecAgent~\cite{zhan2024injecagent} & Safety & $17$ & $1,054$ & Public APIs & \xmark & \cmark & 2024-02 \\ 
        StableToolBench~\cite{guo2024stabletoolbench} & Stable & $16,464$ & $126,486$ & ToolBench2 & \cmark & \cmark & 2024-03 \\ 
        m\&m's~\cite{ma2024m} & Multi-modal & $33$ & $4,427$ &  Public Models & \cmark & \cmark & 2024-03 \\  
        GeoLLM-QA~\cite{singh2024evaluating} & Remote Sensing & $117$ & $1,000$ &  Public Models & \cmark & \cmark & 2024-04 \\  
        ToolLens~\cite{qu2024towards} & Tool Retrieval & $464$ & $18,770$ &  ToolBench2 & \cmark & \cmark & 2024-05\\
        SoAyBench~\cite{wang2024solution} & Academic Seeking & $7$ & $792$ &  AMiner & \cmark & \cmark & 2024-05\\
        ToolSandbox	~\cite{lu2024toolsandbox} & Conversation & $34$ & $1,032$ &  Rapid API & \cmark & \cmark & 2024-08\\
        CToolEval	~\cite{guo2024ctooleval} & Chinese & $398$ & $6,816$ &  Public Apps & \cmark & \cmark & 2024-08\\
        \hline 
    \end{tabular}
    }
    \label{table:benchmark}
\end{table*}

\subsubsection{Information Integration Methods}
\label{subsubsec:response_integration}
Most methodologies opt to incorporate the output of tools into the context as input to LLMs, thereby enabling the LLMs to craft a superior reply based on the information provided by the tool~\cite{shen2024hugginggpt,wang2023tpe,qian2023creator}. 
However, due to the limited context length of LLMs, some tool outputs cannot be directly fed into them. 
Consequently, various methods have emerged to address this issue.
For example, RestGPT~\cite{song2023restgpt} simplifies the lengthy results using the pre-created schema, which is a documentation that elaborates on the examples, format, and possible errors.
ToolLLaMA~\cite{qin2023toolllm} resorts to truncation, cutting the output to fit within the length constraints, which potentially loses the required information to solve the user query. 
Conversely, ReCOMP~\cite{xu2023recomp} develops a compressor to condense lengthy information into a more succinct format, which keeps only the most useful information.
ConAgents~\cite{shi2024learning} proposes a schema-free method, enabling the observing agent to dynamically generate a function adaptive to extracting the target output following the instruction.
And some studies suggest that refining the response generated by LLMs using the tool feedback is more effective than generating the response after invoking the tool~\cite{jacovi2023comprehensive,nathani2023maf,gou2023critic}.

\paratitle{Remark.}
In conclusion, direct insertion methods embed tool outputs directly into the generated response. 
These approaches are straightforward but are only suitable for simple tool outputs. 
Conversely, information integration methods allow LLMs to process tool results to generate responses. 
These methods are more powerful and can provide better responses, enhancing user experience.
However, future work should consider how to address issues related to overly lengthy tool outputs and the inclusion of multiple other modalities.
Meanwhile, some studies highlight that LLMs can generate biased or harmful content and potentially leak sensitive information~\cite{yao2024survey,cui2024risk,das2024security}. By incorporating external tools, whether through direct insertion methods or information integration methods, the generated responses are influenced by the tool results, which can help mitigate some of the biases and harmful content originating from the LLM itself.
Despite these benefits, the introduction of external tools necessitates rigorous validation of tool outputs to prevent adversarial attacks. Without adequate validation, attackers could manipulate tool results, causing LLMs to generate harmful or malicious responses.
Future work should focus on enhancing the ability of LLMs to detect harmful information within tool outputs and develop effective filtering mechanisms to prevent the generation of harmful content.

\section{Benchmarks, Toolkits, and Evaluation}\label{sec:benchmarks and evaluation}

In this section, we systematically summarize and categorize the benchmarks, toolkits, and evaluation methods that are tailored specifically to the various stages of tool learning. This provides a structured overview of the evaluation protocols and toolkits used to validate the effectiveness of tool learning methods, aiming to make it more convenient for readers to engage with and implement tool learning.

\subsection{Benchmarks}
\label{subsec:benchmarks}
With the advancement of research in tool learning, a considerable number of benchmarks have been developed and made available. 
In our survey, we compile a selection of $33$ popular benchmarks~\footnote{Given the growing interest in tool learning,  this survey may not encompass all benchmarks. We welcome suggestions to expand this list.}, as shown in Table~\ref{table:benchmark}.
Each benchmark evaluates distinct facets of tool learning, offering significant contributions to their respective fields.
We categorize these benchmarks into two principal classes: general benchmarks and other benchmarks.

\paratitle{General Benchmarks.}
Given the current uncertainty regarding the capacity of LLMs to effectively utilize tools, a large number of benchmarks have been established to evaluate the tool learning proficiency of LLMs.
As tool learning comprises four distinct stages, existing benchmarks focus on evaluating the capabilities of LLMs at different stages.
For instance, MetaTool~\cite{huang2023metatool} and WTU-Eval~\cite{ning2024wtu} benchmarks are designed to assess whether LLMs can recognize the necessity of using tools and appropriately select the most suitable tool to fulfill user requirements. This assessment particularly focuses on the stages of task planning and tool selection.
On the other hand, APIBench~\cite{patil2023gorilla}, ToolBench1~\cite{xu2023tool}, API-BLEND~\cite{basu2024api}, and Seal-Tools~\cite{wu2024seal} concentrate on the abilities of LLMs to accurately choose the right tool and configure the correct parameters for its invocation, which correspond to the tool selection and tool calling stages, respectively.
Additionally, RestBench~\cite{song2023restgpt}, TaskBench~\cite{shen2023taskbench}, T-Eval~\cite{chen2023t}, and UltraTool~\cite{huang2024planning} extend their focus to include task planning, tool selection, and tool calling, covering three of the four stages. 
Subsequent studies such as API-Bank~\cite{li2023api}, ToolBench2~\cite{qin2023toolllm}, and ToolEyes~\cite{ye2024tooleyes} have provided a more comprehensive evaluation of the tool usage capabilities of LLMs, spanning all four stages of tool learning.
Notably, ToolBench2 has constructed the existing largest tool learning dataset, comprising 16,464 tools and 126,486 instances.
However, the tools included in these benchmarks often suffer from quality issues, such as being inaccessible or non-functional. Additionally, the queries in these benchmarks are typically generated by LLMs, which may not accurately reflect the true needs of users. In contrast, GTA~\cite{wang2024gta}, ShortcutsBench~\cite{shen2024shortcutsbench}, and AppWorld~\cite{trivedi2024appworld} address these limitations by collecting real-world tools and generating queries driven by actual user needs.

\paratitle{Other Benchmarks.}
In addition to general benchmarks, there are also benchmarks specifically designed for particular tasks.
For example, ToolQA~\cite{zhuang2024toolqa} focuses on enhancing the question-answering capabilities of LLMs through the use of external tools, which has developed a dataset comprising questions that LLMs can only answer with the assistance of these external tools.
ToolTalk~\cite{farn2023tooltalk} and ToolSandbox~\cite{lu2024toolsandbox} concentrate on the ability of LLMs to utilize tools within multi-turn dialogues.
VIoT~\cite{zhong2023viotgpt} focuses on the capability of using Viot tools with LLMs.
RoTBench~\cite{ye2024rotbench}, ToolSword~\cite{ye2024toolsword} and ToolEmu~\cite{ruan2023identifying} are benchmarks that emphasize the robustness and safety issues in tool learning. 
These benchmarks highlight the need to improve the robustness and safety of LLMs in tool learning applications.
MLLM-Tool~\cite{wang2024tool} and m\&m's~\cite{ma2024m} extend tool learning into the multi-modal domain, assessing tool usage capabilities of LLMs in multi-modal contexts. 
Meanwhile, StableToolBench~\cite{guo2024stabletoolbench} advocates for the creation of a large-scale and stable benchmark for tool learning. 
SciToolBench~\cite{ma2024sciagent} introduces a novel task named tool-augmented scientific reasoning, expanding the frontier of tool learning with LLMs applications.
GeoLLM-QA~\cite{singh2024evaluating} is designed to capture complex remote sensing workflows where LLMs handle complex data structures, nuanced reasoning, and interactions with dynamic user interfaces.
Moreover, ToolLens~\cite{qu2024towards}, acknowledging that user queries in the real world are often concise yet have ambiguous and complex intent, has created a benchmark focused on the tool retrieval stage.
Building on this, SoayBench~\cite{wang2024solution} introduces a specialized focus on academic information seeking, providing a comprehensive benchmark dataset along with the SOAYEval evaluation method. Lastly, CToolEval~\cite{guo2024ctooleval} extends the concept of tool learning to Chinese societal applications. The CToolEval benchmark is crafted to evaluate the performance of LLMs within the unique context of Chinese societal challenges, emphasizing the applicability and relevance of LLMs in this domain.

\subsection{Toolkits}\label{subsubsec:toolkit}
Recently, several open-sourced libraries and toolkits for achieving tool learning have been proposed.

LangChain is a framework for developing applications powered by large language models. It enables the creation of complex workflows by allowing LLMs to interact with APIs, databases, and other systems. LangChain is ideal for building intelligent assistants and automated systems. It can be accessed at the link: \url{github.com/langchain-ai/langchain}.

Auto-GPT is an open-source application designed to enable large language models to autonomously perform complex tasks with minimal human input. It allows models to break down goals into subtasks and execute them sequentially, including accessing the internet and interacting with APIs. Auto-GPT is ideal for projects requiring autonomous operation, such as automated content creation and research. It can be accessed at the link: \url{github.com/Significant-Gravitas/Auto-GPT}.

BabyAGI is an open-source framework for creating autonomous AI agents that manage and execute tasks with minimal oversight. Its modular design allows developers to extend capabilities by integrating additional tools or APIs, making it ideal for flexible and scalable automation. It can be accessed at the link: \url{github.com/yoheinakajima/babyagi}.

BMTools~\cite{qin2023tool} is an open-source repository designed to enhance LLMs by integrating them with various tools and providing a community-driven platform for developing and sharing these tools. The repository allows for easy plugin creation through simple Python functions and supports the integration of external ChatGPT plugins, making it a powerful resource for extending model capabilities. It can be accessed at the link: \url{github.com/openbmb/BMTools}.

WebCPM~\cite{qin2023webcpm} is a framework designed for developing Chinese long-form question-answering applications using interactive web search. It allows large language models to simulate human-like web browsing behaviors to retrieve and synthesize information from various sources. WebCPM excels in creating sophisticated workflows where LLMs interact with search engines, extract relevant data, and generate comprehensive answers. It can be accessed at the link: \url{github.com/WebCPM/WebCPM}.

\subsection{Evaluation}
\label{subsec:evaluation}

In this section, we will introduce the evaluation methods corresponding to the four stages of tool learning.

\paratitle{Task Planning.} 
The task planning capabilities of LLMs can be evaluated in several ways. Firstly, it is crucial to assess whether LLMs correctly identify if a given query requires a external tool, measuring the accuracy of tool usage awareness~\cite{huang2023metatool,huang2024planning}. 
Next, the effectiveness of the proposed task planning in addressing the query should be evaluated, using metrics like the pass rate provided by ChatGPT~\cite{qin2023toolllm} or human evaluations~\cite{song2023restgpt}. 
Furthermore, the precision of the plan generated by LLMs can be quantitatively analyzed by comparing it to the gold solution, ensuring its alignment and accuracy~\cite{song2023restgpt,chen2023t,qin2023toolllm}.

\paratitle{Tool Selection.}
Existing works employ several metrics to evaluate the effectiveness of tool selection from different perspectives, including Recall, NDCG, and COMP. 

Recall@$K$~\cite{zhu2004recall} is measured by calculating the proportion of selected top-$K$ tools that are present in the set of ground-truth tools:
\[
\text{Recall}\text{@$K$} = \frac{1}{|\mathcal{Q}|} \sum_{q=1}^{|\mathcal{Q}|} \frac{|T_{q}^{K} \cap T_{q}^{*}|}{|T_{q}^{*}|},
\]
where $\mathcal{Q}$ is the set of queries, $T_{q}^{*}$ is the set of relevant tools for the query $q$, and $T_{q}^{K}$ is the top-$K$  tools for the query $q$ selected by the model.

NDCG@$K$~\cite{jarvelin2002cumulated} metric not only considers the proportion of positive tools but also takes into account their positions within the list:
\[
\text{DCG}_q\text{@$K$} = \sum_{i=1}^{K} \frac{2^{g_i}-1}{\log_2{(i+1)}},
\]
\[
\text{NDCG}\text{@$K$} = \frac{1}{|\mathcal{Q}|} \sum_{q=1}^{|\mathcal{Q}|} \frac{\text{DCG}_q\text{@$K$}}{\text{IDCG}_q\text{@$K$}},
\]
where $g_i$ is the graded relevance sore for the $i$-th selected tool, and $\text{IDCG}_q\text{@$K$}$ denotes ideal discounted cumulative gain at the rank position $k$.

COMP@$K$~\cite{qu2024towards} is designed to measure whether the top-$K$ selected tools form a complete set with respect to the ground-truth set:
\[
\text{COMP}\text{@$K$} = \frac{1}{|\mathcal{Q}|} \sum_{q=1}^{|\mathcal{Q}|} \mathbb{I}(\Phi_q \subseteq \Psi^K_q),
\]
where $\Phi_q$ denotes the set of ground-truth tools for query $q$, $\Psi^K_q$ represents the top-$K$ tools retrieved for query $q$, and $\mathbb{I}(\cdot)$ is an indicator function that returns 1 if the retrieval results include all ground-truth tools within the top-$K$ results for query $q$, and 0 otherwise.

\paratitle{Tool Calling.} 
In the stage of tool calling, LLMs are required to generate requests for tool calling in a specified format.
The effectiveness of LLMs in executing tool calling functions can be assessed by evaluating whether the parameters input by LLMs are consistent with the stipulations delineated in the tool documentation~\cite{chen2023t,ye2024tooleyes,huang2024planning}.
This assessment entails verifying whether the parameters provided match those required by the specific tool, including confirming if all required parameters are included and whether the output parameters meet the required range and format.

\paratitle{Response Generation.}
The ultimate goal of tool learning is to enhance the capability of LLMs to effectively address downstream tasks. 
Consequently, the effectiveness of tool utilization is often evaluated based on the performance in solving these downstream tasks~\cite{tang2023toolalpaca,ye2024tooleyes}.
This necessitates that the LLMs consolidate information gathered throughout the entire process, providing a direct response to the user query.
The quality of the final response can be assessed using metrics such as BLEU score~\cite{papineni-etal-2002-bleu}, ROUGE-L~\cite{lin2004rouge}, exact match~\cite{blackwell2009cem}, F1~\cite{basu2024api}, and other relevant indicators.

\section{Challenges and Future Directions}\label{sec:challenge}

In this section, we identify current challenges in tool learning with LLMs and propose some promising directions for future research.

\subsection{High Latency in Tool Learning}
\label{subsec:high latency}
In the reasoning process, LLMs often struggle with high latency and low throughput~\cite{miao2023towards}, challenges that become more pronounced when integrating tool learning.
For example, even simple queries using ChatGPT with plugins can take 5 seconds to resolve, significantly diminishing the user experience compared to faster search engines.
It is essential to explore ways to reduce latency, such as enhancing the awareness of LLMs in tool utilization, enabling them to better assess when the use of tools is genuinely necessary.
Additionally, maintaining the simplicity and responsiveness of tools is crucial.
Overloading a single tool with too many features should be avoided to maintain efficiency and effectiveness.

\subsection{Rigorous and Comprehensive Evaluation}
\label{subsec:Rigorous and Comprehensive Evaluation}
Although current research demonstrates considerable advancements in tool learning with LLMs, evidenced by empirical studies across various applications, there remains a notable gap in establishing solid quantitative metrics to evaluate and understand how effectively LLMs utilize tools. 
Additionally, while numerous strategies have been suggested to enhance the tool learning capabilities of LLMs, a thorough comparative evaluation of these approaches is still missing.
For instance, while human evaluation is capable of accurately reflecting human preferences, it is associated with significant costs and exhibits issues with repeatability, lacking in universal applicability.
While the automated evaluation method, ToolEval~\cite{qin2023toolllm}, has enhanced the efficiency and reproducibility of assessments, it does not necessarily reflect the genuine preference of users.
There is a need for a rigorous and comprehensive evaluation framework that considers efficiency, precision, cost, and practicality holistically. 
Specifically, this framework should provide independent assessments and attribution analysis for improvements at different stages, clearly delineating their specific contributions to the final response. 
This could involve defining new evaluation metrics and constructing assessment environments that simulate the complexity of the real world.

\subsection{Comprehensive and Accessible Tools}
\label{subsec:Comprehensive and Accessible Tools}
While existing efforts have predominantly focused on leveraging tools to enhance the capabilities of LLMs, the quality of these tools critically impacts the performance of tool learning~\cite{wang2024tools}. 
The majority of current tools are aggregated from existing datasets or public APIs, which imposes limitations on their accessibility and comprehensiveness. 
Moreover, existing datasets only contains a limited set of tools, which is unable to cover a diverse range of user queries~\cite{lyu2023gitagent}.
Such constraints curtail the practical applicability and depth of tool learning. 
Additionally, current work acquires tools from different sources such as Public APIs~\cite{tang2023toolalpaca}, RESTful APIs~\cite{song2023restgpt}, Rapid APIs~\cite{qin2023toolllm}, Hugging Face~\cite{patil2023gorilla,shen2024hugginggpt,shen2023taskbench} or the OpenAI plugin list~\cite{huang2023metatool}.
The diverse origins of these tools result in a variance in the format of descriptions, which hampers the development of a unified framework for tool learning. 
There is a pressing need to develop and compile a more comprehensive and easily accessible tool set.
Given the substantial overhead associated with manually creating tools, a viable approach is to employ LLMs for the mass automatic construction of tool set~\cite{cai2023large,wang2024trove}. 
Furthermore, the tool set should encompass a wider range of fields, offering a diverse array of functionalities to meet the specific needs of various domains. 
We posit that a comprehensive and accessible tool set will significantly accelerate the advancement of tool learning.

\subsection{Safe and Robust Tool Learning}
\label{subsec:Safe and Robust Tool Learning}
Current research predominantly emphasizes the capabilities of LLMs in utilizing tools within well-structured environments, yet it overlooks the inevitable presence of noise and emerging safety considerations relevant to real-world applications. 
Upon deploying tool learning with LLMs into practical scenarios, safety issues and noise become unavoidable, necessitating a thoughtful approach to defend against these potential attacks. 
Ye et al. (2024)~\cite{ye2024rotbench} introduce five levels of noise~(Clean, Slight, Medium, Heavy, and Union) to evaluate the robustness of LLMs in tool learning. 
Their findings indicate a significant degradation in performance, revealing that even the most advanced model, GPT-4, exhibits poor resistance to interference. 
Furthermore, Ye et al. (2024)~\cite{ye2024toolsword} devise six safety scenarios to evaluate the safety of LLMs in tool learning, uncovering a pronounced deficiency in safety awareness among LLMs, rendering them incapable of identifying safety issues within tool learning contexts.
With the extensive deployment of tool learning systems across various industries, the imperative for robust security measures has significantly intensified. 
This necessitates a profound investigation and the introduction of innovative methodologies to fortify the security and resilience of these systems. 
Concurrently, in anticipation of emergent attack vectors, it is essential to perpetually refine and update security strategies to align with the rapidly evolving technological landscape.

\subsection{Unified Tool Learning Framework}
\label{subsec:Unified Tool Learning Framework}

As discussed in Section \ref{sec:how tool learning}, the process of tool learning can be categorized into four distinct stages. 
However, prevailing research predominantly concentrates on only one of these stages for specific problems, leading to a fragmented approach and a lack of standardization. This poses significant challenges to scalability and generality in practical scenarios. It is imperative to explore and develop a comprehensive solution that encompasses task planning, tool selection, tool invocation, and response generation within a singular, unified tool learning framework.

\subsection{Real-Word Benchmark for Tool Learning}
\label{subsec:Real-Word Benchmark for Tool Learning}
Despite the substantial volume of work already conducted in the field of tool learning, the majority of queries in existing benchmarks are generated by LLMs rather than originating from real-world user queries. 
These synthesized queries may not accurately reflect genuine human interests and the manner in which users conduct searches. 
To date, there has been no publication of a tool learning dataset that encompasses authentic interactions between users and tool-augmented LLMs.
The release of such a dataset, along with the establishment of a corresponding benchmark, is believed to significantly advance the development of tool learning.

\subsection{Tool Learning with Multi-Modal}
\label{subsec:Tool Learning with Multi-Modal}
While numerous studies have focused on bridging the LLMs with external tools to broaden the application scenarios, the majority of existing work on LLMs in tool learning has been confined to text-based queries. 
This limitation potentially leads to ambiguous interpretations of the true user intent. 
LLMs are poised to enhance understanding of user intent through the integration of visual and auditory information. 
The increasing use of multi-modal data, such as images, audio, 3D, and video, opens up significant opportunities for further development. 
This encompasses exploring the capabilities of multi-modal LLMs in tool use and the combination of multi-modal tools to generate superior responses. 
Several pioneering research projects have explored this area. 
For example, Wang et al. (2024)~\cite{wang2024tool} propose the MLLM-Tool, a system that incorporates open-source LLMs and multi-modal encoders, enabling the learned LLMs to be aware of multi-modal input instructions and subsequently select the correctly matched tool. 
Despite these initial efforts, the exploration of tool learning with multi-modal inputs has not been extensively studied. 
A comprehensive understanding of the capabilities of multi-modal LLMs in tool use is crucial for advancing the field.

\section{Conclusion}\label{sec:conclusion}
In this paper, with reviewing more than 150 papers, we present a comprehensive survey of tool learning with LLMs. 
We begin the survey with a brief introduction to the concepts of 'tool' and 'tool learning,' providing beginners with a foundational overview and essential background knowledge.  
Then we elucidate the benefits of tool integration and tool learning paradigm, detailing six specific aspects to underscore why tool learning is crucial for LLMs.
Moreover, to provide a more detailed introduction to how to conduct tool learning, we break down the tool learning process into four distinct phases: task planning, tool selection, tool calling, and response generation.
Each phase is discussed in depth, integrating the latest research advancements to provide a thorough understanding of each step.
Additionally, we summarize and categorize existing benchmarks and evaluation methods specific to these stages of tool learning, offering a structured overview of evaluation protocols.
Finally, we highlight some potential challenges and identify future directions for research within this evolving field. 
We hope this survey can provide a comprehensive and invaluable resource for researchers and developers keen on navigating the burgeoning domain of tool learning with LLMs, thereby paving the way for future research endeavors.
\begin{acknowledgement}
This work was funded by the National Key R\&D Program of China (2023YFA1008704), the National Natural Science Foundation of China (No. 62377044), Beijing Key Laboratory of Big Data Management and Analysis Methods, Major Innovation \& Planning Interdisciplinary Platform for the ``Double-First Class'' Initiative, funds for building world-class universities (disciplines) of Renmin University of China, and PCC@RUC.
The authors would like to extend their sincere gratitude to Yankai Lin for his constructive feedback throughout the development of this work.
\end{acknowledgement}

\begin{competinginterest}
The authors declare that they have no competing interests or financial conflicts to disclose.
\end{competinginterest}

\bibliographystyle{fcs}
\bibliography{ref}
\vspace{-.5cm}
\begin{biography}{changle}
{Changle Qu} is currently pursuing the Ph.D. degree at Gaoling School of Artificial Intelligence, Renmin University of China. His current research interests mainly include tool learning with large language models and information retrieval. 
\end{biography}
\begin{biography}{sunhao}
{Sunhao Dai} is a Ph.D. candidate at Gaoling School of Artificial Intelligence, Renmin University of China. His current research interests lie in recommender systems and information retrieval. He has published several papers in top-tier conferences such as KDD, SIGIR, ICDE, CIKM, and RecSys.
\end{biography}

\begin{biography}{xiaochi}
{Xiaochi Wei} received Ph.D. degree from Beijing Institute of Technology in 2018, under the supervision of Prof. Heyan Huang. He visited National University of Singapore from 2015 to 2016, under the supervision of Prof. Tat-Seng Chua. He is a Senior R\&D Engineer in Baidu Inc.. His research interests include question answering, multi-media information retrieval, and recommender systems. He has served as PC member in severals conferences, e.g., AAAI, IJCAI, ACL, and EMNLP.
\end{biography}

\begin{biography}{hengyi}
{Hengyi Cai} received Ph.D. degree from Institute of Computing Technology, Chinese Academy of Sciences (Outstanding Graduate) in 2021. He joined JD's doctoral management trainee program in the summer of 2021. Previously, he was a research intern at Baidu's Search Science Team in 2020, under the supervision of Dr. Dawei Yin. His research interests include dialogue system, question answering and information retrieval. He served or is serving as PC member for top-tire conference including ACL, EMNLP, KDD, NeurIPS and SIGIR.
\end{biography}

\begin{biography}{shuaiqiang}
{Shuaiqiang Wang}
received the BSc and Ph.D. degrees in computer science from Shandong University, in 2004 and 2009, respectively. He is currently a principle algorithm engineer with Baidu Inc.. Previously, he was a research scientist with JD.com. Before that, he was an Assistant Professor with the University of Manchester in the U.K. and the University of Jyvaskyla in Finland. served as Senior PC Member of IJCAI, and PC Member of WWW, SIGIR and WSDM in recent years. He is broadly interested in several research areas including information retrieval, recommender systems and data mining.
\end{biography}

\begin{biography}{dawei}
{Dawei Yin}
received Ph.D. degree from Lehigh University in 2013. He is senior director of engineering with Baidu inc.. He is managing the search science team with Baidu. Previously, he was senior director, managing the recommendation engineering team with JD.com between 2016 and 2019. Prior to JD.com, he was senior research manager with Yahoo Labs, leading relevance science team and in charge of Core Search Relevance of Yahoo Search. His research interests include data mining, applied machine learning, information retrieval and recommender system. He published more than 100 research papers in premium conferences and journals, and was the recipients of WSDM 2016 Best Paper Award, KDD 2016 Best Paper Award, WSDM 2018 Best Student Paper Award.
\end{biography}

\begin{biography}{junxu} 
{jun xu} is a professor with the Gaoling School of Artificial Intelligence, Renmin University of China. His research interests focus on learning to rank and semantic matching in web search. He served or is serving as SPC for SIGIR, WWW, and AAAI, editorial board member for JASIST, and associate editor for ACM TOIS. He has won the Test of Time Award Honorable Mention in SIGIR (2019), Best Paper Award in AIRS (2010) and Best Paper Runner-up in CIKM (2017).
\end{biography}

\begin{biography}{jirongwen}
{Ji-rong Wen} 
is a professor of the Renmin University of China (RUC). He is also the dean of the School of Information and executive dean of the Gaoling School of Artificial Intelligence with RUC. His main research interests include information retrieval, data mining, and machine learning. He was a senior researcher and group manager of the Web Search and Mining Group with Microsoft Research Asia (MSRA).
\end{biography}

\end{document}